\documentclass[10pt]{article} 

\usepackage[accepted]{tmlr}

\input{includes}

\usepackage{amsmath,amsfonts,bm}

\def\eqref#1{equation~\ref{#1}}

\def\1{\bm{1}}

\DeclareMathAlphabet{\mathsfit}{\encodingdefault}{\sfdefault}{m}{sl}
\SetMathAlphabet{\mathsfit}{bold}{\encodingdefault}{\sfdefault}{bx}{n}

\DeclareMathOperator{\sign}{sign}

\usepackage{hyperref}
\usepackage{url}

\title{Domain Invariant Adversarial Learning}

\author{\name Matan Levi \email matanle@post.bgu.ac.il \\
      \addr Department of Computer Science\\
      Ben-Gurion University of the Negev
      \AND
      \name Idan Attias \email idanatti@post.bgu.ac.il \\
      \addr Department of Computer Science \\
      Ben-Gurion University of the Negev
      \AND
      \name Aryeh Kontorovich \email karyeh@bgu.ac.il\\
      \addr Department of Computer Science \\
      Ben-Gurion University of the Negev
      }

\def\month{09}  
\def\year{2022} 
\def\openreview{\url{https://openreview.net/forum?id=U8uJAUMzj9}} 

\begin{document}

\maketitle

\begin{abstract}
The phenomenon of adversarial examples illustrates one of the most basic vulnerabilities of deep neural networks. Among the variety of techniques introduced to surmount this inherent weakness, adversarial training has emerged as the most effective strategy for learning robust models. Typically, this is achieved by balancing robust and natural objectives. In this work, we aim to further optimize the trade-off between robust and standard accuracy by enforcing a domain-invariant feature representation. We present a new adversarial training method, \textit{Domain Invariant Adversarial Learning} (DIAL), which learns a feature representation that is both robust and domain invariant. DIAL uses a variant of Domain Adversarial Neural Network (DANN) on the natural domain and its corresponding adversarial domain. In the case where the source domain consists of natural examples and the target domain is the adversarially perturbed examples, our method learns a feature representation constrained not to discriminate between the natural and adversarial examples, and can therefore achieve a more robust representation. DIAL is a generic and modular technique that can be easily incorporated into any adversarial training method. Our experiments indicate that incorporating DIAL in the adversarial training process improves both robustness and standard accuracy.

\end{abstract}

\section{Introduction}
Deep learning models have achieved impressive success on a 
wide
range of challenging tasks. However, their performance was shown to be brittle 
in the face of
\textit{adversarial examples}: small,
imperceptible
perturbations in the input that 
drastically alter the classification
\citep{carlini2017adversarial, carlini2017towards, goodfellow2014explaining, kurakin2016adversarial, moosavi2016deepfool, szegedy2013intriguing, tramer2017ensemble, dong2018boosting, tabacof2016exploring, xie2019improving, rony2019decoupling}.
The problem of
designing reliable robust models has gained significant attention in the arms race against adversarial examples.
Adversarial training \citep{szegedy2013intriguing, goodfellow2014explaining, madry2017towards, zhang2019theoretically} has been proposed as one of the most effective approaches to defend against such examples, and can be described as solving the following min-max optimization problem:
\blfootnote{Our source code is available at \url{https://github.com/matanle51/DIAL}}

\begin{center}
$\min_{\theta}\mathbb{E}_{(x,y)\sim \calD}\sqprn{\max_{x':\norm{x'-x}_p\leq \epsilon}L\paren{x',y;\theta}},$
\end{center}

where $x'$ is the $\epsilon$-bounded perturbation in the $\ell_p$ norm
and $L$ is the loss function.
Different unrestricted attacks methods were also suggested, such as adversarial deformation, rotations, translation and more \citep{brown2018unrestricted, engstrom2018rotation, xiao2018spatially, alaifari2018adef, gilmer2018motivating}.

The resulting min-max optimization problem can be hard to solve in general. Nevertheless, in the context of $\epsilon$-bounded perturbations, the problem is often tractable in practice. The inner maximization is usually approximated by generating adversarial examples using projected gradient descent (PGD) \citep{kurakin2016atscale, madry2017towards}. A PGD adversary starts with randomly initialized perturbation and iteratively adjust the perturbation while projecting it back into the $\epsilon$-ball:

\begin{center}
    $x_{t+1}=\Pi_{\mathbb{B}_\epsilon{(x_0)}}\paren{x_{t} + \alpha\cdot \sign(\nabla_{x_{t}}L(G(x_{t}),y))},$
\end{center}

where $x_0$ is the natural example (with or without random noise), and $\Pi_{\mathbb{B}_\epsilon{(x)}}$ is the projection operator onto the $\epsilon$-ball, $G$ is the network, and $\alpha$ is the perturbation step size.  As was shown by \citet{athalye2018obfuscated}, PGD-based adversarial training was one of the few defenses that were not broken under strong attacks.

That said, the gap between robust and natural accuracy remains large
for many tasks such as CIFAR-10 \citep{krizhevsky2009learning} and ImageNet \citep{deng2009imagenet}. Generally speaking, \citet{tsipras2018robustness} suggested that robustness may be at odds with natural accuracy, and usually the trade-off is inherent. Nevertheless, a growing body of work aimed to improve the standard PGD-based adversarial training introduced by \citet{madry2017towards} in various ways such as improved adversarial loss functions and regularization techniques \citep{kannan2018adversarial, wang2019improving, zhang2019theoretically}, semi-supervised approaches\citep{carmon2019unlabeled, uesato2019labels, zhai2019adversarially}, adversarial perturbations on model weights \citep{wu2020adversarial}, utilizing out of distribution data \citep{lee2021removing}
and many others. 
We refer to related work
for a more extensive literature review.

\paragraph{Our contribution.} In this work, we  propose a novel approach to regulating
the tradeoff between robustness and natural accuracy. In contrast to the aforementioned works, our method enhances adversarial training by enforcing a feature representation that is invariant across the natural and adversarial domains. We incorporate the idea of Domain-Adversarial Neural Networks (DANN)~\citep{ ganin2015unsupervised, ganin2016domain} directly into the adversarial training process. DANN is a representation learning approach for domain adaptation, designed to ensure that predictions are made based on invariant feature representation that cannot discriminate between source and target domains. This technique is modular and can be easily incorporated into any standard adversarial training algorithm. Intuitively, the tasks of adversarial training and of domain-invariant representation have a similar goal: given a source (natural) domain $X$ and a target (adversarial) domain $X'$, we hope to achieve $g(X) \approx g(X')$, where $g$ is a feature representation function (i.e., neural network). As we present in section \ref{theory}, our work is also theoretically motivated by the domain adaptation generalization bounds.

In a comprehensive battery of experiments on MNIST~\citep{lecun1998gradient}, SVHN~\citep{netzer2011reading}, CIFAR-10~\citep{krizhevsky2009learning} and CIFAR-100~\citep{krizhevsky2009learning} datasets,
we demonstrate that by enforcing domain-invariant representation learning using DANN simultaneously with adversarial training, we gain a significant and consistent improvement in both robustness and natural accuracy compared to other state-of-the-art adversarial training methods, under Auto-Attack~\citep{croce2020reliable} and various strong PGD~\citep{madry2017towards}, and CW~\citep{carlini2017towards} adversaries in white-box and black-box settings. 
Additionally, we evaluate our method using unforeseen ``natural'' corruptions~\citep{hendrycks2018benchmarking}, unforeseen adversaries (e.g., $\ell_{1}$, $\ell_{2}$), transfer  learning, and perform ablation studies.
Finally, we offer a novel score function for  quantifying the robust-natural accuracy trade-off.

\begin{figure}[H]
  \centering
  \includegraphics[width=0.5\textwidth]{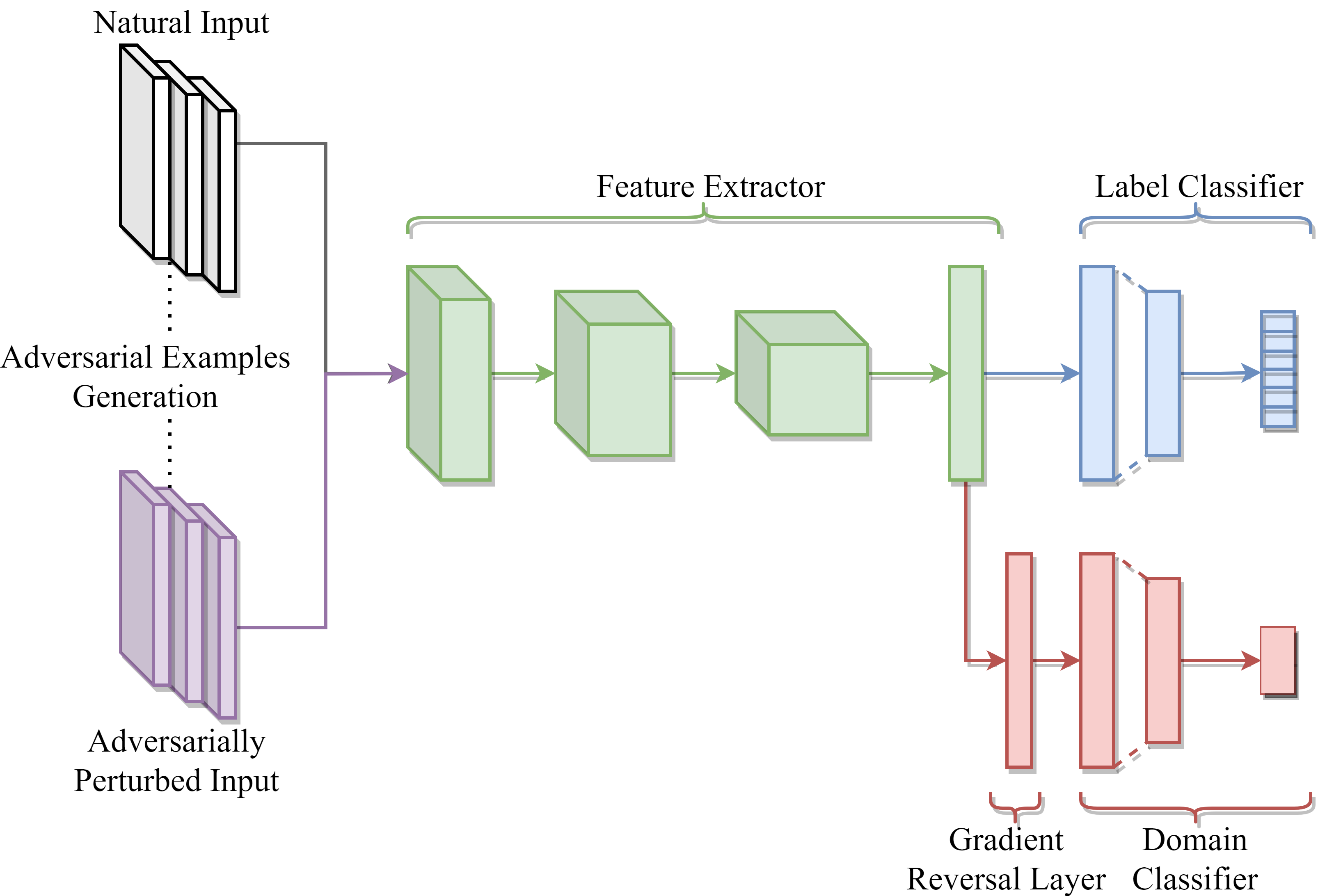}
  \caption{Illustration of the proposed architecture to enforce domain invariant representation. The feature extractor and label classifier form the a regular DNN architecture that can be used for the main natural task. The domain classifier is incorporated alongside the label classifier. The reversal gradient layer multiplies the gradient by a negative number during the back-propagation.}
  \label{dann}
\end{figure}

\section{Related work}
\subsection{Defense methods}
A variety of theoretically principled~\citep{raghunathan2018certified, sinha2017certifiable, raghunathan2018semidefinite, wong2018scaling, wong2018provable, gowal2018effectiveness} and empirical defense approaches \citep{bai2021recent} were proposed to enhance robustness since the discovery of adversarial examples. 
Theoretically principled methods focus on certifying robustness to adversarial perturbations under a given norm, using variety of techniques such as randomized smoothing \citep{cohen2019certified}. However, empirical defence methods in general, and in particular adversarial training, still yield preferable results.

Among the empirical defence techniques, we can find: \textit{adversarial regularization}---adding various regularization terms to the loss functions to enhance robustness (e.g., encouraging logits for clean and adversarial examples to be similar) \citep{kurakin2016atscale, madry2017towards, zhang2019theoretically, wang2019improving, kannan2018adversarial, jin2022enhancing}, \textit{curriculum-based adversarial training}---taking incremental approach when learning PGD adversaries in the goal of improving generalization on clean data while still preserving robustness (e.g., gradually increasing the number of PGD iterations to avoid overfitting the adversarial examples) \citep{cai2018curriculum, zhang2020attacks, wang2019convergence}, \textit{ensemble adversarial training}---where clean data is augmented with adversarial
examples generated from different target models instead of
a single model \citep{tramer2017ensemble, pang2019improving, yang2020dverge}, \textit{adversarial training with adaptive attack budget}---where we change the perturbation budget to prevent over-confident predictions and achieve better exploration of the manifold \citep{ding2018mma, cheng2020cat}, \textit{semi-supervised and unsupervised adversarial training}---several methods theoretically and empirically demonstrated how unlabeled data can reduce the sample complexity gap between
standard training and adversarial training  \citep{carmon2019unlabeled, uesato2019labels, zhai2019adversarially}, \textit{robust self and pre-training}---other works integrate self-supervised pretraining tasks such as Selfie, Rotation and Jigsaw together with adversarial examples \citep{jiang2020robust, chen2020adversarial}, \textit{efficient adversarial training}---due to the high cost of adversarial training, efficient methods aim to keep the favorable performance of adversarial training while reducing the computational and time costs \citep{shafahi2019adversarial, wong2020fast, andriushchenko2020understanding, zhang2019you}, and many other techniques such as adversarial training based on feature scatter \citep{zhang2019defense}, adversarially robust distillation \citep{goldblum2020adversarially}, hypersphere embedding \citep{pang2020boosting}, and augmenting adversarial examples by interpolation \citep{lee2020adversarial}. In an additional research direction, researchers suggested to add new dedicated building blocks to the network architecture for improved robustness \citep{xie2019intriguing, xie2019feature, liu2020towards}. \citet{liu2020towards} hypothesised that different adversaries belong to different domains, and suggested gated batch normalization which is trained with multiple perturbation types. \cite{guo2020meets} focused on searching robust architectures against adversarial examples. Others works presented improved robustness by combining data augmentation techniques and generated data \citep{rebuffi2021fixing, rebuffi2021data}, where the latest is also the current state-of-the-art in robustness. 

Our work belongs to the the family of adversarial regularization techniques, for which we elaborate on common and best performing methods, and highlight the differences compared to our method. 

\citet{madry2017towards} proposed a technique, commonly referred to as Adversarial Training (AT), to minimize the cross entropy loss on adversarial examples generated by PGD (without using the natural examples). \citet{zhang2019theoretically}
suggested to decompose the prediction error for adversarial examples as the sum of the natural error and boundary error, and provided differentiable upper bounds on both terms. Motivated by this decomposition, they suggested a technique called TRADES that uses the Kullback-Leibler (KL) divergence as a regularization term that will push the decision boundary away from the data. They do so by applying the KL-divergence on the logits of clean examples and their adversarial counterparts. 
\citet{wang2019improving} suggested that misclassified examples have a significant impact on 
final robustness, and proposed a technique called MART that differentiate between correctly classified and miss-classified examples during training by weighting the KL-divergence between the clean and adversarial logits using the probability of the classifier on the correct label. 

Another area of research aims at revealing the connection between the loss weight landscape
and adversarial training \citep{prabhu2019understanding, yu2018interpreting, wu2020adversarial}. Specifically,~\citet{wu2020adversarial} identified 
a correlation between the flatness of weight loss landscape and robust generalization gap. They proposed
the
Adversarial Weight Perturbation (AWP) mechanism that is integrated into existing adversarial training methods and generates adversarial perturbations on both the inputs and the network weights. More recently, this approach was formalized from a theoretical standpoint by~\citet{tsai2021formalizing}. However, this method forms a double-perturbation mechanism that perturbs both inputs and weights,
which may incur a significant increase in calculation overhead. We demonstrate how DIAL improves results also when combined with AWP, named $\DIAL_{\awp}$.
In Section \ref{related-loss-func} we elaborate about the loss functions of the different compared methods.

A related approach to ours, called ATDA,
was presented by \citet{song2018improving}.
They proposed to add several constrains to the loss function in order to enforce domain adaptation: correlation alignment and maximum mean discrepancy~\citep{borgwardt2006integrating, sun2016deep}. While the objective is similar, using ideas from domain adaptation for learning better representation, we address it in two different ways. Our method fundamentally differs from~\citet{song2018improving} since we do not enforce domain adaptation by adding specific constrains to the loss function. Instead, we let the network learn the domain invariant representation directly during the optimization process, as suggested by~\citet{ganin2015unsupervised,ganin2016domain}. Moreover,~\citet{song2018improving} focused mainly of Fast Gradient Sign Method (FGSM) attack, which is a one step variant of PGD attack. We empirically demonstrate the superiority of our method in Section~\ref{experiments}. 
In a concurrent work, \citet{qian2021improving} utilized the idea of exploiting local and global data information, and suggested to generate the adversarial examples by attacking an additional domain classifier.

\subsection{Theoretical analysis of robust generalization}
Several works investigated the sample complexity requires the ensure adversarial generalization compared to the non-adversarial counterpart.
\citet{schmidt2018adversarially} has shown that there exists a distribution (mixture of Gaussians) where ensuring robust generalization necessarily requires more data than standard learning. This has been furthered investigated in a distribution-free models via the Rademacher complexity, VC dimension, and fat-shattering dimension ~\citep{yin2019rademacher,attias2019improved,khim2018adversarial,awasthi2020rademacher,cullina2018pac,montasser2019vc,tsai2021formalizing,attias2022characterization,attias2022adversarially} and additional settings~\citep{diochnos2018adversarial,carmon2019unlabeled}.


\section{Domain Invariant Adversarial Learning approach}
In this section, we introduce our Domain Invariant Adversarial Learning (DIAL) approach for adversarial training. The source domain is the natural dataset, and the target domain is generated using adversarial attack on the natural domain. We aim to learn a model that has low error on the source (natural) task (e.g., classification) while ensuring that the internal representation cannot discriminate between the natural and adversarial domains. 
In this way,
we enforce additional regularization on the feature representation, which 
enhances robustness.


\subsection{The benefits of invariant representation to adversarial examples}
The motivation behind the proposed method is to enforce an invariant feature representation to adversarial perturbations. Given a natural example $x$ and its adversarial counterpart $x'$, if the domain classifier manages to distinguish between 
them, this means
that the perturbation has induced a 
significant
difference in the feature representation. We impose an additional loss on the natural and adversarial domains in order to discourage this behavior. 

\begin{figure}[ht]
\centering
  \subfigure[Mean difference comparison]{\includegraphics[width=0.47\textwidth]{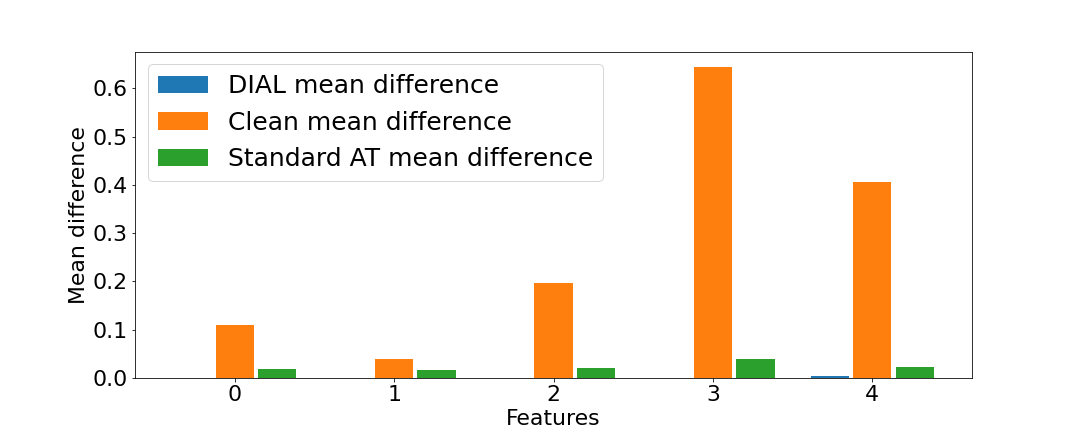}}
  \subfigure[Standard deviation difference comparison]{\includegraphics[width=0.47\textwidth]{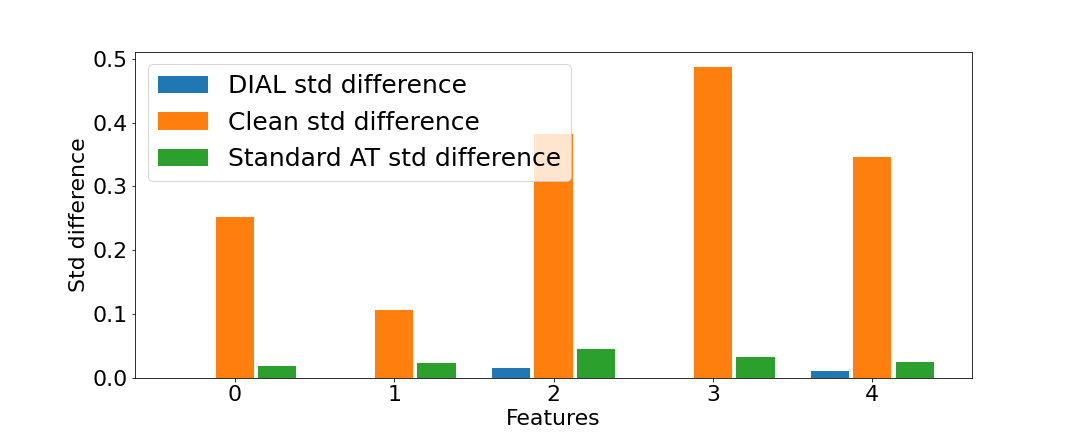}}
    \caption{We visualize the (a) Mean and (b) standard deviation (std) differences comparison between three models: (1) Naturally trained model (without adversarial training), named Clean. (2) Model trained using standard adversarial training, named standard AT, and (3) Model trained using our method, DIAL.
    We visualize five random features from the features layer. Each bar represent the difference between the means/std of the natural examples and the mean/std of their corresponding adversarial examples on this same feature.}
  \label{feat_layer_stat}
  \label{pgd_cifar10}
\end{figure}

We demonstrate that the feature representation layer does not discriminate between natural and adversarial examples, namely $G_f(x;\theta_f)\approx G_f(x';\theta_f)$. Figure  \ref{feat_layer_stat} presents the scaled mean and standard deviation (std)
of the
absolute differences between the natural examples from test and their 
corresponding 
adversarial 
examples on different features from the feature representation layer.
Smaller differences in the mean and std
imply a higher domain invariance
--- and indeed, DIAL achieves
near-zero differences almost across the board.
Moreover, DIAL's feature-level invariance almost consistently 
outperforms
the naturally trained model (model trained without adversarial training), and the model trained using standard adversarial training
techniques~\citep{madry2017towards}. We provide additional features visualizations in Appendix \ref{additional_viz}. 

Recently, other communities also discovered the benefits of adopting analogous architectures to DANN, such as the contrastive learning community which used similar architecture to improve representation learning \citep{dangovski2021equivariant, wang2021residual}

\subsection{Model architecture and regularized loss function}
\label{dial-loss-section}
Let us define the notation for our domain invariant robust architecture and loss. Let $G_f(\cdot;\theta_f)$ be the feature extractor neural network with parameters $\theta_f$. Let $G_y(\cdot;\theta_y)$ be the label classifier with parameters $\theta_y$, and let $G_d(\cdot;\theta_d)$ be the domain classifier with parameters $\theta_d$. 
That is, $G_y(G_f(\cdot;\theta_f);\theta_y)$ is essentially the standard model (e.g., wide residual network~\citep{zagoruyko2016wide}), while in addition, we have a domain classification layer to enforce a domain invariant on the feature representation. An illustration of the architecture is presented in Figure~\ref{dann}.

Given a training set $\{{(x_i,y_i)}\}_{i=1}^{n}$, the natural loss is defined as:
\begin{center}
$\mathcal{L}_{\nat}^y=\frac{1}{n}\sum_{i=1}^{n}{\text{CE}(G_y(G_f(x_i;\theta_f);\theta_y),y_i)}$. 
\end{center}

We consider two basic
forms
of
the robust loss. One is the standard cross-entropy (CE) loss between the predicted probabilities and the actual label, which we refer to later as $\DIAL_{\ce}$. The second is the Kullback-Leibler (KL) divergence between the adversarial and natural model outputs (logits), i.e., the class probabilities between the natural examples and their adversarial counterparts, which we refer to as $\DIAL_{\kl}$.

\begin{center}
$\mathcal{L}_{\robloss}^{\ce}= \frac{1}{n}\sum_{i=1}^{n}{\text{CE}(G_y(G_f(x'_i;\theta_f);\theta_y),y_i)}$,

$\mathcal{L}_{\robloss}^{\kl}= \frac{1}{n}\sum_{i=1}^{n}{\text{KL}(G_y(G_f(x'_i;\theta_f); \theta_y)  \parallel G_y(G_f(x_i;\theta_f); \theta_y))}$.
\end{center}

where $\{{(x'_{i},y_i)}\}_{i=1}^{n}$ are the generated corresponding adversarial examples. Next, we define source domain label $d_i$ as 0 (for natural examples) and target domain label $d_i^{'}$ as 1 (for adversarial examples). Then, the natural and adversarial domain losses are defined as:

\begin{center}$
\mathcal{L}_{\nat}^d=\frac{1}{n}\sum_{i=1}^{n}{\text{CE}(G_d(G_f(x_i;\theta_f);\theta_d),d_i)}$,

$\mathcal{L}_{\adv}^d=\frac{1}{n}\sum_{i=1}^{n}{\text{CE}(G_d(G_f(x_i^{'};\theta_f);\theta_d),d'_i)}$.
\end{center}

We can now define the full domain invariant robust loss:

\begin{center}
$\DIAL_{\ce} = \mathcal{L}_{\nat}^y + \lambda\mathcal{L}_{\robloss}^{\ce} -r(\mathcal{L}_{\nat}^d + \mathcal{L}_{\adv}^d)$,

$\DIAL_{\kl} = \mathcal{L}_{\nat}^y + \lambda\mathcal{L}_{\robloss}^{\kl} -r(\mathcal{L}_{\nat}^d + \mathcal{L}_{\adv}^d)$.
\end{center}

The goal is to \textit{minimize} the loss on the natural and adversarial classification while \textit{maximizing} the loss for the domains. 
The {\em reversal-ratio} hyper-parameter 
$r$
is inserted into the network layers as a gradient reversal layer~\citep{ ganin2015unsupervised, ganin2016domain}  that leaves the input unchanged
during forward propagation and reverses the gradient by multiplying it with a negative scalar
during the back-propagation. 
The reversal-ratio parameter is initialized to a small value and is gradually increased to $r$, as the main objective converges.
This enforces a domain-invariant representation as the training progress: a larger value enforces a higher fidelity to the domain.
\hide{
During the next phase, the standard back-propagation algorithm is modified as follows: instead of updating with a positive multiple of the gradient, scalar $-r$
is used.
\ak{OLD:
This parameter is tuned during the training phase to facilitate the main task converge at the beginning of the training, and enforce a domain invariant representation as the training progress. 
}}
A comprehensive algorithm description can be found in Appendix \ref{algo-appendix}.


\subsection{Theoretical Analysis through the Lens of Generalization Bounds}
\label{theory}
Our method refers to the natural and adversarial examples as two distinct domains, where the source domain consists of natural examples and the target domain is the adversarially perturbed examples. Given this assumption, we can adapt the generalization bounds from \citet{mansour2009domain} to theoretically justify our approach.

\paragraph{Preliminaries.}
Let $H$ be a set of functions mapping $X$ to $Y$ and let $\mathcal{L}:Y \times Y\rightarrow\mathbb{R}_+$ be a loss function over $Y$.
The $\mathcal{L}_Q(f, g)$ loss functional is defined as the expected loss for any
two functions $f,g : X \rightarrow Y$ in $H$, and any distribution $Q$ over $X$:
\begin{center} 
$\mathcal{L}_Q(f,g) = \mathbb{E}_{x \sim Q}[L(f(x),g(x))]$.
\end{center}

We also denote $f_Q: X \rightarrow Y$ as the 
target function on examples drawn from $Q$. 
That is, the error of a function $h\in H$ is defined as 
$\mathcal{L}_Q(h,f_Q) = \mathbb{E}_{x \sim Q}[L(h(x),f_Q(x))]$.


We define the  discrepancy distance $\disc_L$ between
two distributions $Q_1$ and $Q_2$ over $X$ by:
\begin{center} 
$\disc_L(Q_1,Q_2)=\max_{h,h'\in H} |\mathcal{L}_{Q_1}(h,h')-\mathcal{L}_{Q_2}(h,h')|$.
\end{center}

\paragraph{Reduction to domain adaptation.} Given source (natural) distribution $D$, we define the target (adversarial) distribution $D_{\text{adv}}$, such that every pair $(x,y)$ in the support of $D$ is mapped to $(z,y)$, where $z\in \mathbb{B}_{\epsilon}(x)$ and $D_{\text{adv}}(z,y)=D(x,y)$.

Let $h_{D}^{\star}=\arg \min_{h\in H}\mathcal{L}_D(h,f_{D})$ be the optimal natural function, and let $h_{D_{\text{adv}}}^{\star}=\arg \min_{h\in H}\mathcal{L}_{D_{\text{adv}}}(h,f_{D_{\text{adv}}})$ be the optimal robust function.
Additionally, we assume that the labeling function $f_{D_{\text{adv}}}$ is the same as $f_D$.

Under these assumptions, we reduce our problem to a domain adaptation one, which consists of selecting a hypothesis $h \in H$ with a small expected
loss according to the target distribution.

We can now adapt Theorem 8 in \citep{mansour2009domain} to our case and bound the adversarial loss by:

\begin{equation}
\begin{aligned}
\label{bound}
\mathcal{L}_{D_{\text{adv}}}(h,f_{D_{\text{adv}}})
&\leq
\overbrace{ \mathcal{L}_{D_{\text{adv}}}(h_{D_{\text{adv}}}^{\star}, f_{D_{\text{adv}}})}^{\text{adv}}
+
\overbrace{\mathcal{\mathcal{L}}_{D}(h,h_{D}^{\star})}^{\text{natural}}
\\
&+
\underbrace{\mathcal{L}_{D}(h^{\star}_{D},h^{\star}_{D_\text{adv}})}_{\text{trade-off}}
+\underbrace{\disc_{L}(D_{\text{adv}}, D)}_{\text{discrepancy}}.
\end{aligned}
\end{equation}

The bound on the adversarial loss consists of four terms: \textit{first term} - approximation error of $H$ for adversarial distribution. \textit{second term} - estimation error on natural examples for the output function on the true distribution (compared to the optimal $h \in H$). \textit{third term} - trade-off error (depends on $H$, $D$, $D_{adv}$, and not the algorithm output) that represents the fraction of points for which the optimal $h \in H$ on adversarial examples are wrong, compared to the optimal $h \in H$ on clean examples. \textit{forth term} - the discrepancy distance between $D$ and $D_{adv}$.

Therefore, the adversarial loss bounded in \eqref{bound} depends, in addition to the natural and adversarial losses, on the discrepancy distance, $\disc_{L}$, between the two distributions. The discrepancy distance will decrease as the two distributions $D$ and $D_{\text{adv}}$ will be similar to one another.

Our adversarial training method, DIAL, minimizes the adversarial and natural losses, and simultaneously learns an invariant feature representation between the adversarial and natural domains. This representation ensures that adversarial examples and their natural counterparts will be invariant to the domain they were taken from, and therefore makes the two distributions $D$ and $D_{\text{adv}}$ similar to each other.
\textit{Altogether, DIAL also minimizes the discrepancy distance, which in turn leads to minimizing the upper bound on the adversarial error}.












\subsection{Related work loss function comparison}
\label{related-loss-func}
In Table \ref{loss-func-comaprison} we further illustrate the loss functions of the different methods. 
For $\TRADES_{\awp}$, the process involves running TRADES loss function, and then running weight perturbation as defined in equation (10) in \citep{wu2020adversarial}.
These loss functions can be compared to our proposed loss functions $\DIAL_{\ce}$ and $\DIAL_{\kl}$ which are detailed in section \ref{dial-loss-section}, where the main difference lies in the new natural and adversarial domain losses. We colored in red the unique terms of our method.

\begin{table}[ht]
  \caption{Loss function comparison. Let (x,y) be the natural example and its corresponding label. Let x' be the adversarial example generated from x. $\text{CE}$ refers to the cross-entropy loss function, $\text{KL}$ refers to the KL-divergence loss function, and BCE is the boosted cross-entropy loss function. $\mathcal{L}_{CORAL}$ and $\mathcal{L}_{MMD}$ correspond to the correlation alignment and maximum mean discrepancy~\citep{borgwardt2006integrating, sun2016deep}, respectively. $\mathcal{L}_{margin}$ minimize the intra-class variations and maximize the inter-class variations \citep{song2018improving}.
  $\lambda$ is a hyper parameters to control the ratio between different losses, and $r$ is the reversal ratio hyper parameter. 
  Let $G_f(\cdot;\theta_f)$ be the feature extractor neural network with parameters $\theta_f$. Let $G_y(\cdot;\theta_y)$ be the label classifier with parameters $\theta_y$, and let $G_d(\cdot;\theta_d)$ be the domain classifier with parameters $\theta_d$. That is, $G(\cdot;\theta) = G_y(G_f(\cdot;\theta_f);\theta_y)$ is essentially the standard model definition (e.g., wide residual network).We define source domain label $d$ as 0 (for natural examples) and target domain label $d^{'}$ as 1 (for adversarial examples).
  For convenience, we present the loss function on a single example.}
  \vskip 0.1in
  \label{loss-func-comaprison}
  \centering
  \small
  \begin{tabular}{l|c}
    \toprule
    Method & Loss function \\
    \midrule
    AT & $\text{CE}(G(x';\theta),y)$ \\
    \midrule
    TRADES & $ \text{CE}(G(x;\theta),y) + \lambda \cdot \text{KL}(G(x';\theta) \parallel G(x;\theta))$ \\
    \midrule
    MART & $ 
    \text{BCE}(G(x';\theta),y) + \lambda \cdot \text{KL}(G(x';\theta) \parallel G(x;\theta))\cdot(1- G(x;\theta)_y)$ \\
    \midrule
    ATDA & $ \text{CE}(G(x';\theta),y) + \text{CE}(G(x;\theta),y) + \mathcal{L}_{CORAL} + \mathcal{L}_{MMD} + \mathcal{L}_{margin}$ \\
    \midrule
    \midrule
    $\DIAL_{\ce}$ & $\text{CE}(G(x;\theta),y) + \lambda\cdot\text{CE}(G(x';\theta),y) \textcolor{red}{-r(\text{CE}(G_d(G_f(x;\theta_f);\theta_d),d) + \text{CE}(G_d(G_f(x';\theta_f);\theta_d),d'))}$ \\
    \midrule
    $\DIAL_{\kl}$ & $\text{CE}(G(x;\theta),y) + \lambda\cdot\text{KL}(G(x';\theta) \parallel G(x;\theta)) \textcolor{red}{-r(\text{CE}(G_d(G_f(x;\theta_f);\theta_d),d) + \text{CE}(G_d(G_f(x';\theta_f);\theta_d),d'))}$ \\
    \bottomrule
  \end{tabular}
\end{table}

\section{Experiments} \label{experiments}
In this section we conduct comprehensive experiments to emphasise the effectiveness of DIAL, including evaluations under white-box and black-box settings, robustness to unforeseen adversaries, robustness to unforeseen corruptions, transfer learning, and ablation studies. Finally, we present a new measurement to test the balance between robustness and natural accuracy, which we named $F_1$-robust score.

\subsection{A case study on SVHN and CIFAR-100}
In the first part of our analysis, we conduct a case study experiment on two benchmark datasets: SVHN \citep{netzer2011reading} and CIFAR-100 \cite{krizhevsky2009learning}. We follow common experiment settings as in \cite{rice2020overfitting, wu2020adversarial}. We used the PreAct ResNet-18 \citep{he2016identity} architecture on which we integrate a domain classification layer. The adversarial training is done using 10-step PGD adversary with perturbation size of 0.031 and a step size of 0.003 for SVHN and 0.007 for CIFAR-100. The batch size is 128, weight decay is $7e^{-4}$ and the model is trained for 100 epochs. For SVHN, the initial learinnig rate is set to 0.01 and decays by a factor of 10 after 55, 75 and 90 iteration. For CIFAR-100, the initial learning rate is set to 0.1 and decays by a factor of 10 after 75 and 90 iterations. 
Results are averaged over 3 restarts while omitting one standard deviation (which is smaller than 0.2\% in all experiments). As can be seen by the results in Tables~\ref{black-and_white-svhn} and \ref{black-and_white-cifar100}, DIAL presents consistent improvement in robustness (e.g., 5.75\% improved robustness on SVHN against AA) compared to the standard AT 
while also improving the natural accuracy. More results are presented in Appendix \ref{cifar100-svhn-appendix}.

\begin{table}[!ht]
  \caption{Robustness against white-box, black-box attacks and Auto-Attack (AA) on SVHN. Black-box attacks are generated using naturally trained surrogate model. Natural represents the naturally trained (non-adversarial) model.
  }
  \vskip 0.1in
  \label{black-and_white-svhn}
  \centering
  \small
  \begin{tabular}{l@{\hspace{1\tabcolsep}}c@{\hspace{1\tabcolsep}}c@{\hspace{1\tabcolsep}}c@{\hspace{1\tabcolsep}}c@{\hspace{1\tabcolsep}}c@{\hspace{1\tabcolsep}}c@{\hspace{1\tabcolsep}}c@{\hspace{1\tabcolsep}}c@{\hspace{1\tabcolsep}}c@{\hspace{1\tabcolsep}}c}
    \toprule
    & & \multicolumn{4}{c}{White-box} & \multicolumn{4}{c}{Black-Box}  \\
    \cmidrule(r){3-6} 
    \cmidrule(r){7-10}
    Defense Model & Natural & PGD$^{20}$ & PGD$^{100}$  & PGD$^{1000}$  & CW$^{\infty}$ & PGD$^{20}$ & PGD$^{100}$ & PGD$^{1000}$  & CW$^{\infty}$ & AA \\
    \midrule
    NATURAL & 96.85 & 0 & 0 & 0 & 0 & 0 & 0 & 0 & 0 & 0 \\
    \midrule
    AT & 89.90 & 53.23 & 49.45 & 49.23 & 48.25 & 86.44 & 86.28 & 86.18 & 86.42 & 45.25 \\
    $\DIAL_{\kl}$ (Ours) & 90.66 & \textbf{58.91} & \textbf{55.30} & \textbf{55.11} & \textbf{53.67} & 87.62 & 87.52 & 87.41 & 87.63 & \textbf{51.00} \\
    $\DIAL_{\ce}$ (Ours) & \textbf{92.88} & 55.26  & 50.82 & 50.54 & 49.66 & \textbf{89.12} & \textbf{89.01} & \textbf{88.74} & \textbf{89.10} &  46.52  \\
    \bottomrule
  \end{tabular}
\end{table}

\begin{table}[!ht]
  \caption{Robustness against white-box, black-box attacks and Auto-Attack (AA) on CIFAR100. Black-box attacks are generated using naturally trained surrogate model. Natural represents the naturally trained (non-adversarial) model.
  }
  \vskip 0.1in
  \label{black-and_white-cifar100}
  \centering
  \small
  \begin{tabular}{l@{\hspace{1\tabcolsep}}c@{\hspace{1\tabcolsep}}c@{\hspace{1\tabcolsep}}c@{\hspace{1\tabcolsep}}c@{\hspace{1\tabcolsep}}c@{\hspace{1\tabcolsep}}c@{\hspace{1\tabcolsep}}c@{\hspace{1\tabcolsep}}c@{\hspace{1\tabcolsep}}c@{\hspace{1\tabcolsep}}c}
    \toprule
    & & \multicolumn{4}{c}{White-box} & \multicolumn{4}{c}{Black-Box}  \\
    \cmidrule(r){3-6} 
    \cmidrule(r){7-10}
    Defense Model & Natural & PGD$^{20}$ & PGD$^{100}$  & PGD$^{1000}$  & CW$^{\infty}$ & PGD$^{20}$ & PGD$^{100}$ & PGD$^{1000}$  & CW$^{\infty}$ & AA \\
    \midrule
    NATURAL & 79.30 & 0 & 0 & 0 & 0 & 0 & 0 & 0 & 0 & 0 \\
    \midrule
    AT & 56.73 & 29.57 & 28.45 & 28.39 & 26.6 & 55.52 & 55.29 & 55.26 & 55.40 & 24.12 \\
    $\DIAL_{\kl}$ (Ours) & 58.47 & \textbf{31.19} & \textbf{30.50} & \textbf{30.42} & \textbf{26.91} & 57.16 & 56.81 & 56.80 & 57.00 & \textbf{25.87} \\
    $\DIAL_{\ce}$ (Ours) & \textbf{60.77} & 27.87 & 26.66 & 26.61 & 25.98 & \textbf{59.48} & \textbf{59.06} & \textbf{58.96} & \textbf{59.20} & 23.51  \\
    \bottomrule
  \end{tabular}
\end{table}


\subsection{Performance comparison on CIFAR-10} \label{defence-settings}
In this part, we evaluate the performance of DIAL compared to other well-known methods on CIFAR-10. 
We follow the same experiment setups as in~\cite{madry2017towards, wang2019improving, zhang2019theoretically}. When experiment settings are not identical between tested methods, we choose the most commonly used settings, and apply it to all experiments. This way, we keep the comparison as fair as possible and avoid reporting changes in results which are caused by inconsistent experiment settings \citep{pang2020bag}. To show that our results are not caused because of what is referred to as \textit{obfuscated gradients}~\citep{athalye2018obfuscated}, we evaluate our method with same setup as in our defense model, under strong attacks (e.g., PGD$^{1000}$) in both white-box, black-box settings, Auto-Attack ~\citep{croce2020reliable}, unforeseen "natural" corruptions~\citep{hendrycks2018benchmarking}, and unforeseen adversaries. To make sure that the reported improvements are not caused by \textit{adversarial overfitting}~\citep{rice2020overfitting}, we report best robust results for each method on average of 3 restarts, while omitting one standard deviation (which is smaller than 0.2\% in all experiments). Additional results for CIFAR-10 as well as comprehensive evaluation on MNIST can be found in Appendix \ref{mnist-results} and \ref{additional_res}.

\begin{table}[ht]
  \caption{Robustness against white-box, black-box attacks and Auto-Attack (AA) on CIFAR-10. Black-box attacks are generated using naturally trained surrogate model. Natural represents the naturally trained (non-adversarial) model.
  }
  \vskip 0.1in
  \label{black-and_white-cifar}
  \centering
  \small
  \begin{tabular}{cccccccc@{\hspace{1\tabcolsep}}c}
    \toprule
    & & \multicolumn{3}{c}{White-box} & \multicolumn{3}{c}{Black-Box} \\
    \cmidrule(r){3-5} 
    \cmidrule(r){6-8}
    Defense Model & Natural & PGD$^{20}$ & PGD$^{100}$ & CW$^{\infty}$ & PGD$^{20}$ & PGD$^{100}$ & CW$^{\infty}$ & AA \\
    \midrule
    NATURAL & 95.43 & 0 & 0 & 0 & 0 & 0 & 0 &  0 \\
    \midrule
    TRADES & 84.92 & 56.60 & 55.56 & 54.20 & 84.08 & 83.89 & 83.91 &  53.08 \\
    MART & 83.62 & 58.12 & 56.48 & 53.09 & 82.82 & 82.52 & 82.80 & 51.10 \\
    AT & 85.10 & 56.28 & 54.46 & 53.99 & 84.22 & 84.14 & 83.92 & 51.52 \\
    ATDA & 76.91 & 43.27 & 41.13 & 41.01 & 75.59 & 75.37 & 75.35 & 40.08\\
    $\DIAL_{\kl}$ (Ours) & 85.25 & $\mathbf{58.43}$ & $\mathbf{56.80}$ & $\mathbf{55.00}$ & 84.30 & 84.18 & 84.05 & \textbf{53.75} \\
    $\DIAL_{\ce}$ (Ours)  & $\mathbf{89.59}$ & 54.31 & 51.67 & 52.04 &$ \mathbf{88.60}$ & $\mathbf{88.39}$ & $\mathbf{88.44}$ & 49.85 \\
    \midrule
    $\DIAL_{\awp}$ (Ours) & $\mathbf{85.91}$ & $\mathbf{61.10}$ & $\mathbf{59.86}$ & $\mathbf{57.67}$ & $\mathbf{85.13}$ & $\mathbf{84.93}$ & $\mathbf{85.03}$  & \textbf{56.78} \\
    $\TRADES_{\awp}$ & 85.36 & 59.27 & 59.12 & 57.07 & 84.58 & 84.58 & 84.59 & 56.17 \\
    \bottomrule
  \end{tabular}
\end{table}

\paragraph{CIFAR-10 setup.} We use the wide residual network (WRN-34-10)~\citep{zagoruyko2016wide} architecture. 
Sidelong this architecture, we integrate a domain classification layer. To generate the adversarial domain dataset, we use a perturbation size of $\epsilon=0.031$. We apply 10 of inner maximization iterations with perturbation step size of 0.007. Batch size is set to 128, weight decay is set to $7e^{-4}$, and the model is trained for 100 epochs. Similar to the other methods, the initial learning rate was set to 0.1, and decays by a factor of 10 at iterations 75 and 90. 
See Appendix \ref{cifar10-additional-setup} for additional details.

\begin{table}[ht]
  \caption{Black-box attack using the adversarially trained surrogate models on CIFAR-10.}
  \vskip 0.1in
  \label{black-box-cifar-adv}
  \centering
  \small
  \begin{tabular}{ll|c}
    \toprule
    \cmidrule(r){1-2}
    Surrogate (source) model & Target model & robustness \% \\
    \midrule
    TRADES & $\DIAL_{\ce}$ & $\mathbf{67.77}$ \\
    $\DIAL_{\ce}$ & TRADES & 65.75 \\
    \midrule
    MART & $\DIAL_{\ce}$ & $\mathbf{70.30}$ \\
    $\DIAL_{\ce}$ & MART & 64.91 \\
    \midrule
    AT & $\DIAL_{\ce}$ & $\mathbf{65.32}$ \\
    $\DIAL_{\ce}$ & AT  & 63.54 \\
    \midrule
    ATDA & $\DIAL_{\ce}$ & $\mathbf{66.77}$ \\
    $\DIAL_{\ce}$ & ATDA & 52.56 \\
    \bottomrule
  \end{tabular}
\end{table}

\paragraph{White-box/Black-box robustness.} 
As reported in Table~\ref{black-and_white-cifar} and Appendix~\ref{additional_res}, our method achieves better robustness compared to the other methods. Specifically, in the white-box settings, our method improves robustness over~\citet{madry2017towards} and TRADES by 2\% 
while keeping higher natural accuracy. We also observe better natural accuracy of 1.65\% over MART while also achieving better robustness over all attacks. Moreover, our method presents significant improvement of up to 15\% compared to the the domain invariant method suggested by~\citet{song2018improving} (ATDA).
When incorporating 
AWP, our method improves the results of $\TRADES_{\awp}$ by almost 2\%.
When tested on black-box settings, $\DIAL_{\ce}$ presents a significant improvement of more than 4.4\% over the second-best performing method, and up to 13\%. In Table~\ref{black-box-cifar-adv}, we also present the black-box results when the source model is taken from one of the adversarially trained models. 
In addition to the improvement in black-box robustness, $\DIAL_{\ce}$ also manages to achieve better clean accuracy of more than 4.5\% over the second-best performing method.

\subsubsection{Robustness to Unforeseen Attacks and Corruptions}
\paragraph{Unforeseen Adversaries.} To further demonstrate the effectiveness of our approach, we test our method against various adversaries that were not used during the training process. We attack the model under the white-box settings with $\ell_{2}$-PGD, $\ell_{1}$-PGD, $\ell_{\infty}$-DeepFool and $\ell_{2}$-DeepFool \citep{moosavi2016deepfool} adversaries using Foolbox \citep{rauber2017foolbox}. We applied commonly used attack budget 
with 20 and 50 iterations for PGD and DeepFool, respectively.
Results are presented in Table \ref{unseen-attacks}. As can be seen, our approach  gains an improvement of up to 4.73\% over the second best method under the various attack types and an average improvement of 3.7\% over all threat models.

\begin{table}[ht]
  \caption{Robustness on CIFAR-10 against unseen adversaries under white-box settings.}
  \vskip 0.1in
  \label{unseen-attacks}
  \centering
  \begin{tabular}{c@{\hspace{1.5\tabcolsep}}c@{\hspace{1.5\tabcolsep}}c@{\hspace{1.5\tabcolsep}}c@{\hspace{1.5\tabcolsep}}c@{\hspace{1.5\tabcolsep}}c@{\hspace{1.5\tabcolsep}}c@{\hspace{1.5\tabcolsep}}c}
    \toprule
    Threat Model & Attack Constraints & $\DIAL_{\kl}$ & $\DIAL_{\ce}$ & AT & TRADES & MART & ATDA \\
    \midrule
    \multirow{2}{*}{$\ell_{2}$-PGD} & $\epsilon=0.5$ & 76.05 & \textbf{80.51} & 76.82 & 76.57 & 75.07 & 66.25 \\
    & $\epsilon=0.25$ & 80.98 & \textbf{85.38} & 81.41 & 81.10 & 80.04 & 71.87 \\\midrule
    \multirow{2}{*}{$\ell_{1}$-PGD} & $\epsilon=12$ & 74.84 & \textbf{80.00} & 76.17 & 75.52 & 75.95 & 65.76 \\
    & $\epsilon=7.84$ & 78.69 & \textbf{83.62} & 79.86 & 79.16 & 78.55 & 69.97 \\
    \midrule
    $\ell_{2}$-DeepFool & overshoot=0.02 & 84.53 & \textbf{88.88} & 84.15 & 84.23 & 82.96 & 76.08 \\\midrule
    $\ell_{\infty}$-DeepFool & overshoot=0.02 & 68.43 & \textbf{69.50} & 67.29 & 67.60 & 66.40 & 57.35 \\
    \bottomrule
  \end{tabular}
\end{table}

\begin{figure}[h]
 \centering
  \includegraphics[width=0.4\textwidth]{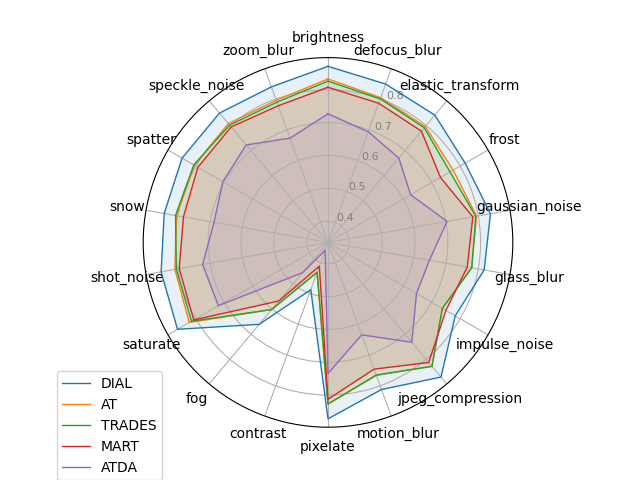}
  \caption{Accuracy comparison over all unforeseen corruptions.}
  \label{corruption}
\end{figure}

\paragraph{Unforeseen Corruptions.}
We further demonstrate that our method consistently holds against unforeseen ``natural'' corruptions, consists of 18 unforeseen diverse corruption types proposed by \citet{hendrycks2018benchmarking} on CIFAR-10, which we refer to as CIFAR10-C. The CIFAR10-C benchmark covers noise, blur, weather, and digital categories. As can be shown in Figure \ref{corruption}, our method gains a significant and consistent improvement over all the other methods. Our method leads to an average improvement of 4.7\% with minimum improvement of 3.5\% and maximum improvement of 5.9\% compared to the second best method over all unforeseen attacks. See Appendix \ref{corruptions-apendix} for the full experiment results.

\subsubsection{Transfer Learning}
Recent works \citep{salman2020adversarially,utrera2020adversarially} suggested that robust models transfer better on standard downstream classification tasks. In Table \ref{transfer-res} we demonstrate the advantage of our method when applied for transfer learning across CIFAR10 and CIFAR100 using the common linear evaluation protocol. see Appendix \ref{transfer-learning-settings} for detailed settings.

\begin{table}[H]
  \caption{Transfer learning results comparison.}
  \vskip 0.1in
  \label{transfer-res}
  \centering
  \small
\begin{tabular}{c|c|c|c}
\toprule

\multicolumn{2}{l}{} & \multicolumn{2}{c}{Target} \\
\cmidrule(r){3-4}
Source & Defence Model & CIFAR10 & CIFAR100 \\
\midrule
\multirow{3}{*}{CIFAR10} & DIAL & \multirow{3}{*}{-} & \textbf{28.57} \\
 & AT &  & 26.95  \\
 & TRADES &  & 25.40  \\
 \midrule
\multirow{3}{*}{CIFAR100} & DIAL & \textbf{73.68} & \multirow{3}{*}{-} \\
 & AT & 71.41 & \\
 & TRADES & 71.42 &  \\
\bottomrule
\end{tabular}
\end{table}

\subsubsection{Modularity and Ablation Studies}

We note that the domain classifier is a modular component that can be integrated into existing models for further improvements. Removing the domain head and related loss components from the different DIAL formulations results in some common adversarial training techniques. For $\DIAL_{\kl}$, removing the domain and related loss components results in the formulation of TRADES. For $\DIAL_{\ce}$, removing the domain and related loss components results in the original formulation of the standard adversarial training, and for $\DIAL_{\awp}$ the removal results in $\TRADES_{\awp}$. Therefore, the ablation studies will demonstrate the effectiveness of combining DIAL on top of different adversarial training methods. 

\begin{figure}[ht]
  \centering
  \includegraphics[width=0.35\textwidth]{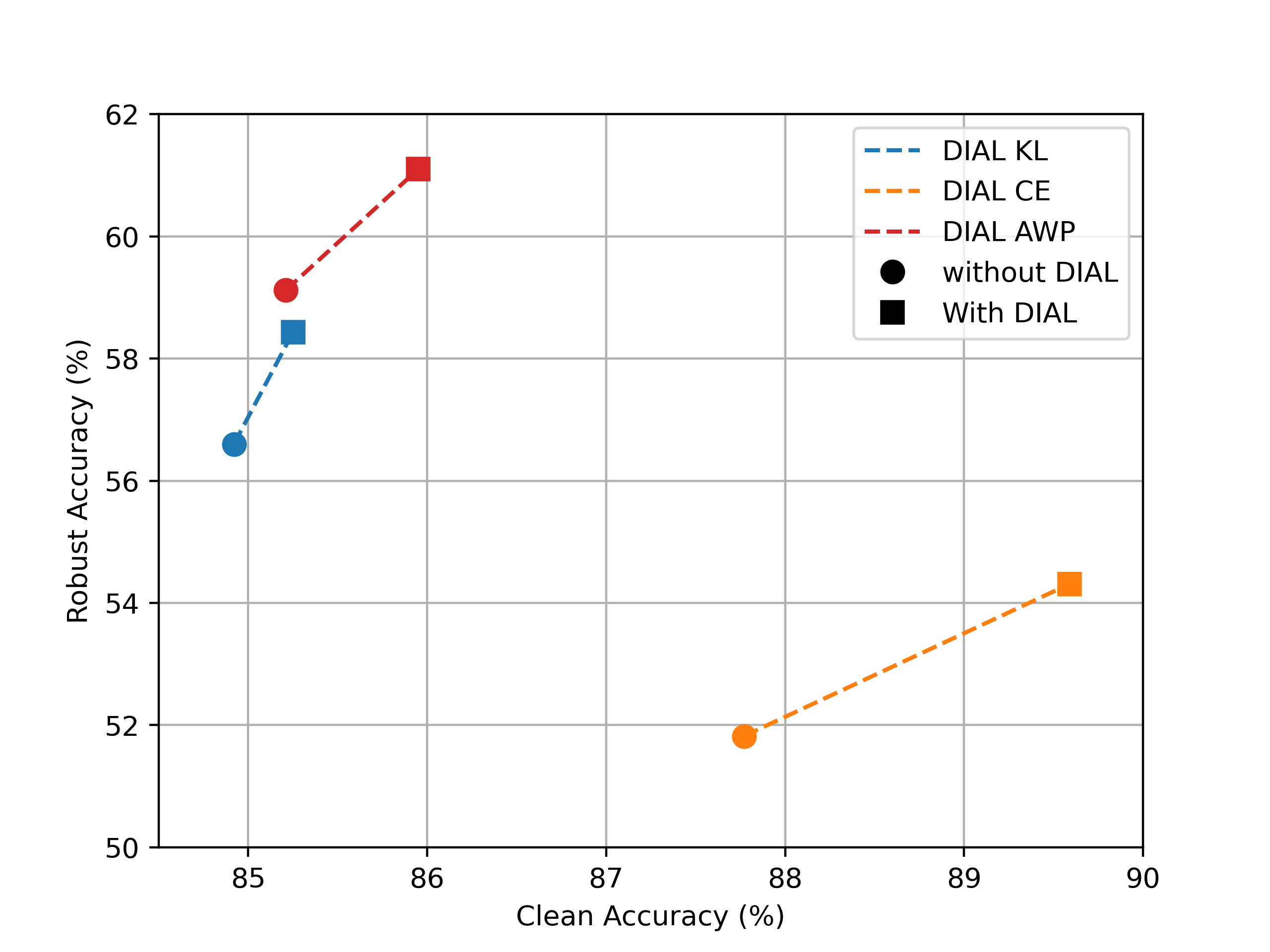}
  \caption{Ablation studies for $\DIAL_{\kl}$, $\DIAL_{\ce}$, and $\DIAL_{\awp}$ on CIFAR-10. Circle represent the robust-natural accuracy without using DIAL, and square represent the robust-natural accuracy when incorporating DIAL.
  }
  \label{ablation}
\end{figure}

We investigate the contribution of the additional domain head component introduced in our method. Experiment configuration are as in \ref{defence-settings}, and robust accuracy is based on white-box PGD$^{20}$ on CIFAR-10 dataset. We remove the domain head from both $\DIAL_{\kl}$, $\DIAL_{\awp}$, and $\DIAL_{\ce}$ (equivalent to $r=0$) and report the natural and robust accuracy. We perform 3 random restarts and omit one standard deviation from the results. Results are presented in Figure \ref{ablation}. All DIAL variants exhibits stable improvements on both natural accuracy and robust accuracy. $\DIAL_{\ce}$, $\DIAL_{\kl}$, and $\DIAL_{\awp}$ present an improvement of 1.82\%, 0.33\%, and 0.55\% on natural accuracy and an improvement of 2.5\%, 1.87\%, and 0.83\% on robust accuracy, respectively. This evaluation empirically demonstrates the benefits of incorporating DIAL on top of different adversarial training techniques.

\subsubsection{Visualizing DIAL}
\begin{figure}[h]
\centering
  \subfigure[\textbf{DIAL} on natural logits]{\includegraphics[width=0.21\textwidth]{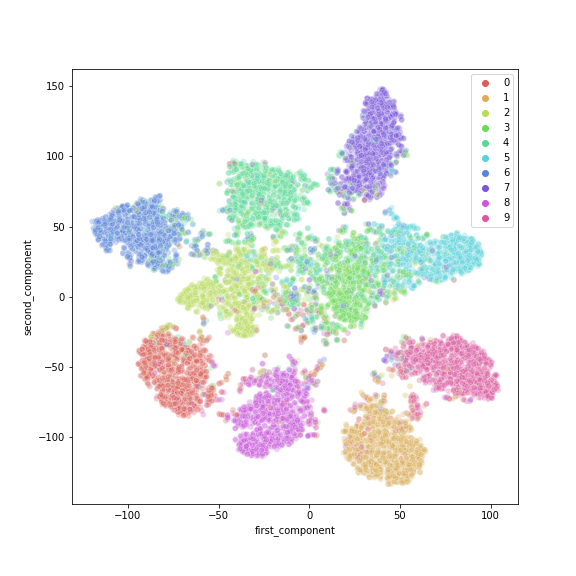}}
  \hspace{0.03\textwidth}
  \subfigure[\textbf{DIAL} on adversarial logits]{\includegraphics[width=0.21\textwidth]{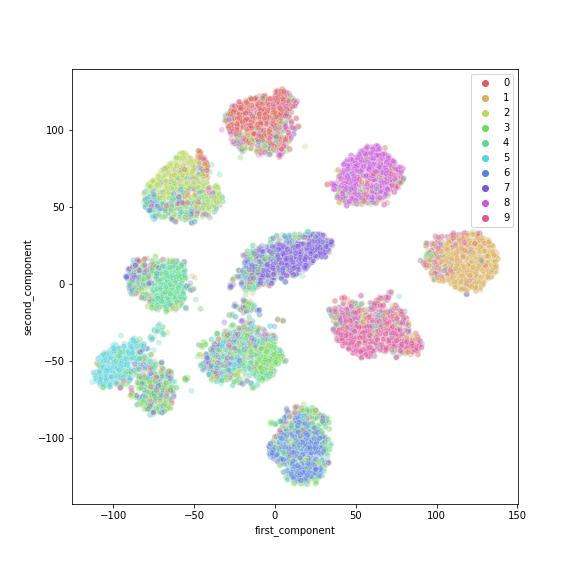}}
  \hspace{0.03\textwidth}
    \subfigure[\textbf{TRADES} on natural logits]{\includegraphics[width=0.21\textwidth]{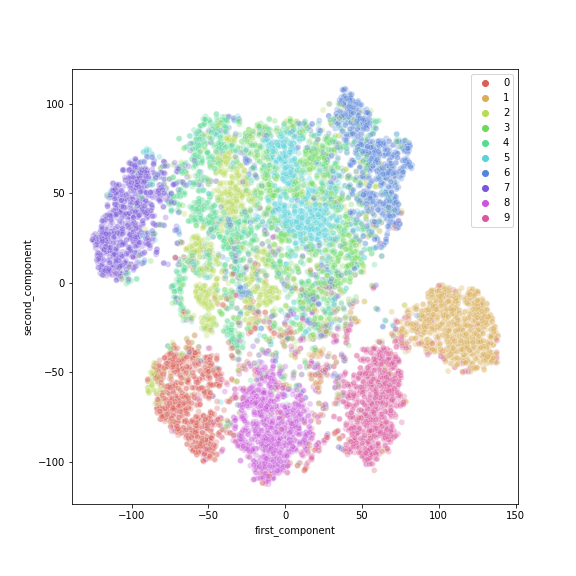}}
    \hspace{0.03\textwidth}
    \subfigure[\textbf{TRADES} on adversarial logits]{\includegraphics[width=0.21\textwidth]{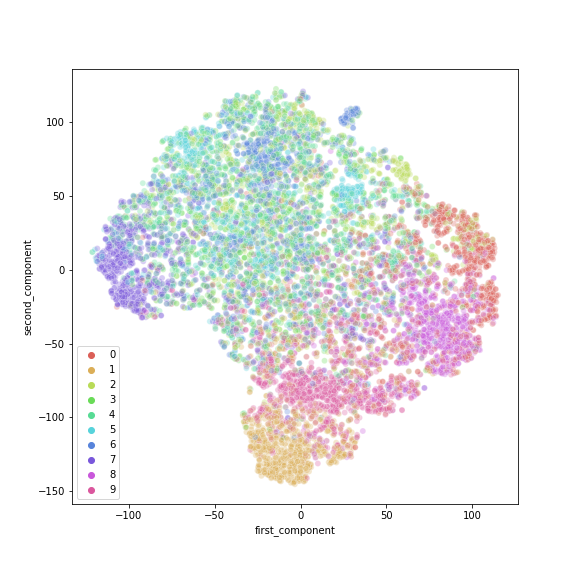}}
  \caption{t-SNE embedding of model output (logits) into two-dimensional space for DIAL and TRADES using the CIFAR-10 natural test data and the corresponding PGD$^{20}$ generated adversarial examples.}
  \label{tsne1}
\end{figure}

To further illustrate the superiority of our method, we visualize the model outputs from the different methods on both natural and adversarial test data.
Figure~\ref{tsne1} shows the embedding received after applying t-SNE ~\citep{van2008visualizing} with two components on the model output for our method and for TRADES. DIAL seems to preserve strong separation between classes on both natural test data and adversarial test data. Additional illustrations for the other methods are attached in Appendix~\ref{additional_viz}. 

\subsection{Balanced measurement for robust-natural accuracy}
One of the goals of our method is to better balance between robust and natural accuracy under a given model. For a balanced metric, we adopt the idea of $F_1$-score, which is the harmonic mean between the precision and recall. However, rather than using precision and recall, we measure the $F_1$-score between robustness and natural accuracy,
using a measure we call
the
\textbf{$\mathbf{F_1}$-robust} score.
\begin{equation}
F_1\text{-robust} = \dfrac{\text{true\_robust}}
{\text{true\_robust}+\frac{1}{2}
(\text{false\_{robust}}+
\text{false\_natural})},
\end{equation}
where $\text{true\_robust}$ are the adversarial examples that were correctly classified, $\text{false\_{robust}}$ are the adversarial examples that were miss-classified, and $\text{false\_natural}$ are the natural examples that were miss-classified.
Results are presented in Table~\ref{f1-robust} and demonstrate that our method achieves the best $F_1$-robust score in both settings, which supports our findings from previous sections.


\begin{table}[ht]
\small
  \caption{$F_1$-robust measurement using PGD$^{20}$ attack in white and black box settings on CIFAR-10.}
  \vskip 0.1in
  \label{f1-robust}
  \centering
  \begin{tabular}{c
  @{\hspace{1.5\tabcolsep}}c @{\hspace{1.5\tabcolsep}}c @{\hspace{1.5\tabcolsep}}c @{\hspace{1.5\tabcolsep}}c
  @{\hspace{1.5\tabcolsep}}c @{\hspace{1.5\tabcolsep}}c @{\hspace{1.5\tabcolsep}}|
  @{\hspace{1.5\tabcolsep}}c
  @{\hspace{1.5\tabcolsep}}c}
    \toprule
     & TRADES & MART & AT & ATDA & $\DIAL_{\ce}$ & $\DIAL_{\kl}$ & $\DIAL_{\awp}$ & $\TRADES_{\awp}$ \\
    \midrule
    White-box & 0.659 & 0.666 & 0.657 & 0.518 & 0.660 & \textbf{0.675} & \textbf{0.698} & 0.682 \\
    Black-box & 0.844 & 0.831 & 0.845 & 0.761 & \textbf{0.890} & 0.847 & \textbf{0.854} & 0.849 \\ 
    \bottomrule
  \end{tabular}
\end{table}

\section{Conclusion and Future Work}
In this paper, we investigated the hypothesis that domain invariant representation can be beneficial for robust learning. With this idea in mind, we proposed a new adversarial learning method, called \textit{Domain Invariant Adversarial Learning} (DIAL) that incorporates domain adversarial neural network into the adversarial training process.
The proposed method, DIAL, is theoretically motivated by the domain adaptation generalization bounds.
DIAL is generic and can be combined with any network architecture and any adversarial training technique in a wide range of tasks. Additionally, since the domain classifier does not require the class labels, we argue that additional unlabeled data can be leveraged in future work. Our evaluation process included strong adversaries, unforeseen adversaries, unforeseen corruptions, transfer learning tasks, and ablation studies. Using the extensive empirical analysis, we demonstrate the significant and consistent improvement obtained by DIAL in both robustness and natural accuracy. 

\section{Broader Impact Statement}
Adversarial examples illustrate a fundamental vulnerability of deep neural networks, and manage to break state-of-the art DNNs in various fields. As these DNNs are deployed in critical systems, such as autonomous vehicles and facial recognition systems, it becomes crucial to build models which are robust against such attacks. These kind of system cannot tolerate attacks that can cost in human lives.
For this reason, we proposed DIAL to improve models' robustness against adversarial attacks.
We hope that it will help in building more secure models for real-world applications.  DIAL 
is comparable 
to the state-of-the-art methods we tested
in terms of training times and other
resources.
That said, this work is not without limitations: adversarial training is still a computationally expensive procedure that requires extra computations compared to standard training, 
with the concomitant environmental costs.
Even though incorporating our method introduced improved standard accuracy, adversarial training still degrades the standard accuracy. 
Moreover, models are trained to be robust using well known threat models such as the bounded $\ell_{p}$ norms. However, once a model is deployed, we cannot control the type of attacks it faces from sophisticated adversaries. Thus, the general problem
is still very far from being fully solved.

\bibliography{refs.bib}

\begin{thebibliography}{96}
\providecommand{\natexlab}[1]{#1}
\providecommand{\url}[1]{\texttt{#1}}
\expandafter\ifx\csname urlstyle\endcsname\relax
  \providecommand{\doi}[1]{doi: #1}\else
  \providecommand{\doi}{doi: \begingroup \urlstyle{rm}\Url}\fi

\bibitem[Alaifari et~al.(2018)Alaifari, Alberti, and
  Gauksson]{alaifari2018adef}
Rima Alaifari, Giovanni~S Alberti, and Tandri Gauksson.
\newblock Adef: an iterative algorithm to construct adversarial deformations.
\newblock \emph{arXiv preprint arXiv:1804.07729}, 2018.

\bibitem[Andriushchenko \& Flammarion(2020)Andriushchenko and
  Flammarion]{andriushchenko2020understanding}
Maksym Andriushchenko and Nicolas Flammarion.
\newblock Understanding and improving fast adversarial training.
\newblock \emph{arXiv preprint arXiv:2007.02617}, 2020.

\bibitem[Athalye et~al.(2018)Athalye, Carlini, and
  Wagner]{athalye2018obfuscated}
Anish Athalye, Nicholas Carlini, and David Wagner.
\newblock Obfuscated gradients give a false sense of security: Circumventing
  defenses to adversarial examples.
\newblock In \emph{International Conference on Machine Learning}, pp.\
  274--283. PMLR, 2018.

\bibitem[Attias \& Hanneke(2022)Attias and Hanneke]{attias2022adversarially}
Idan Attias and Steve Hanneke.
\newblock Adversarially robust learning of real-valued functions.
\newblock \emph{arXiv preprint arXiv:2206.12977}, 2022.

\bibitem[Attias et~al.(2019)Attias, Kontorovich, and
  Mansour]{attias2019improved}
Idan Attias, Aryeh Kontorovich, and Yishay Mansour.
\newblock Improved generalization bounds for robust learning.
\newblock In \emph{Algorithmic Learning Theory}, pp.\  162--183. PMLR, 2019.

\bibitem[Attias et~al.(2022)Attias, Hanneke, and
  Mansour]{attias2022characterization}
Idan Attias, Steve Hanneke, and Yishay Mansour.
\newblock A characterization of semi-supervised adversarially-robust pac
  learnability.
\newblock \emph{arXiv preprint arXiv:2202.05420}, 2022.

\bibitem[Awasthi et~al.(2020)Awasthi, Frank, and Mohri]{awasthi2020rademacher}
Pranjal Awasthi, Natalie Frank, and Mehryar Mohri.
\newblock On the rademacher complexity of linear hypothesis sets.
\newblock \emph{arXiv preprint arXiv:2007.11045}, 2020.

\bibitem[Bai et~al.(2021)Bai, Luo, Zhao, Wen, and Wang]{bai2021recent}
Tao Bai, Jinqi Luo, Jun Zhao, Bihan Wen, and Qian Wang.
\newblock Recent advances in adversarial training for adversarial robustness.
\newblock \emph{arXiv preprint arXiv:2102.01356}, 2021.

\bibitem[Borgwardt et~al.(2006)Borgwardt, Gretton, Rasch, Kriegel,
  Sch{\"o}lkopf, and Smola]{borgwardt2006integrating}
Karsten~M Borgwardt, Arthur Gretton, Malte~J Rasch, Hans-Peter Kriegel,
  Bernhard Sch{\"o}lkopf, and Alex~J Smola.
\newblock Integrating structured biological data by kernel maximum mean
  discrepancy.
\newblock \emph{Bioinformatics}, 22\penalty0 (14):\penalty0 e49--e57, 2006.

\bibitem[Brown et~al.(2018)Brown, Carlini, Zhang, Olsson, Christiano, and
  Goodfellow]{brown2018unrestricted}
Tom~B Brown, Nicholas Carlini, Chiyuan Zhang, Catherine Olsson, Paul
  Christiano, and Ian Goodfellow.
\newblock Unrestricted adversarial examples.
\newblock \emph{arXiv preprint arXiv:1809.08352}, 2018.

\bibitem[Cai et~al.(2018)Cai, Du, Liu, and Song]{cai2018curriculum}
Qi-Zhi Cai, Min Du, Chang Liu, and Dawn Song.
\newblock Curriculum adversarial training.
\newblock \emph{arXiv preprint arXiv:1805.04807}, 2018.

\bibitem[Carlini \& Wagner(2017{\natexlab{a}})Carlini and
  Wagner]{carlini2017adversarial}
Nicholas Carlini and David Wagner.
\newblock Adversarial examples are not easily detected: Bypassing ten detection
  methods.
\newblock In \emph{Proceedings of the 10th ACM Workshop on Artificial
  Intelligence and Security}, pp.\  3--14, 2017{\natexlab{a}}.

\bibitem[Carlini \& Wagner(2017{\natexlab{b}})Carlini and
  Wagner]{carlini2017towards}
Nicholas Carlini and David Wagner.
\newblock Towards evaluating the robustness of neural networks.
\newblock In \emph{2017 ieee symposium on security and privacy (sp)}, pp.\
  39--57. IEEE, 2017{\natexlab{b}}.

\bibitem[Carmon et~al.(2019)Carmon, Raghunathan, Schmidt, Liang, and
  Duchi]{carmon2019unlabeled}
Yair Carmon, Aditi Raghunathan, Ludwig Schmidt, Percy Liang, and John~C Duchi.
\newblock Unlabeled data improves adversarial robustness.
\newblock \emph{arXiv preprint arXiv:1905.13736}, 2019.

\bibitem[Chen et~al.(2020)Chen, Liu, Chang, Cheng, Amini, and
  Wang]{chen2020adversarial}
Tianlong Chen, Sijia Liu, Shiyu Chang, Yu~Cheng, Lisa Amini, and Zhangyang
  Wang.
\newblock Adversarial robustness: From self-supervised pre-training to
  fine-tuning.
\newblock In \emph{Proceedings of the IEEE/CVF Conference on Computer Vision
  and Pattern Recognition}, pp.\  699--708, 2020.

\bibitem[Cheng et~al.(2020)Cheng, Lei, Chen, Dhillon, and Hsieh]{cheng2020cat}
Minhao Cheng, Qi~Lei, Pin-Yu Chen, Inderjit Dhillon, and Cho-Jui Hsieh.
\newblock Cat: Customized adversarial training for improved robustness.
\newblock \emph{arXiv preprint arXiv:2002.06789}, 2020.

\bibitem[Cohen et~al.(2019)Cohen, Rosenfeld, and Kolter]{cohen2019certified}
Jeremy Cohen, Elan Rosenfeld, and Zico Kolter.
\newblock Certified adversarial robustness via randomized smoothing.
\newblock In \emph{International Conference on Machine Learning}, pp.\
  1310--1320. PMLR, 2019.

\bibitem[Croce \& Hein(2020)Croce and Hein]{croce2020reliable}
Francesco Croce and Matthias Hein.
\newblock Reliable evaluation of adversarial robustness with an ensemble of
  diverse parameter-free attacks.
\newblock In \emph{ICML}, 2020.

\bibitem[Cullina et~al.(2018)Cullina, Bhagoji, and Mittal]{cullina2018pac}
Daniel Cullina, Arjun~Nitin Bhagoji, and Prateek Mittal.
\newblock Pac-learning in the presence of adversaries.
\newblock In \emph{Advances in Neural Information Processing Systems}, pp.\
  230--241, 2018.

\bibitem[Dangovski et~al.(2021)Dangovski, Jing, Loh, Han, Srivastava, Cheung,
  Agrawal, and Soljacic]{dangovski2021equivariant}
Rumen Dangovski, Li~Jing, Charlotte Loh, Seungwook Han, Akash Srivastava, Brian
  Cheung, Pulkit Agrawal, and Marin Soljacic.
\newblock Equivariant self-supervised learning: Encouraging equivariance in
  representations.
\newblock In \emph{International Conference on Learning Representations}, 2021.

\bibitem[Deng et~al.(2009)Deng, Dong, Socher, Li, Li, and
  Fei-Fei]{deng2009imagenet}
Jia Deng, Wei Dong, Richard Socher, Li-Jia Li, Kai Li, and Li~Fei-Fei.
\newblock Imagenet: A large-scale hierarchical image database.
\newblock In \emph{2009 IEEE conference on computer vision and pattern
  recognition}, pp.\  248--255. Ieee, 2009.

\bibitem[Ding et~al.(2018)Ding, Sharma, Lui, and Huang]{ding2018mma}
Gavin~Weiguang Ding, Yash Sharma, Kry Yik~Chau Lui, and Ruitong Huang.
\newblock Mma training: Direct input space margin maximization through
  adversarial training.
\newblock \emph{arXiv preprint arXiv:1812.02637}, 2018.

\bibitem[Diochnos et~al.(2018)Diochnos, Mahloujifar, and
  Mahmoody]{diochnos2018adversarial}
Dimitrios Diochnos, Saeed Mahloujifar, and Mohammad Mahmoody.
\newblock Adversarial risk and robustness: General definitions and implications
  for the uniform distribution.
\newblock In \emph{Advances in Neural Information Processing Systems}, pp.\
  10359--10368, 2018.

\bibitem[Dong et~al.(2018)Dong, Liao, Pang, Su, Zhu, Hu, and
  Li]{dong2018boosting}
Yinpeng Dong, Fangzhou Liao, Tianyu Pang, Hang Su, Jun Zhu, Xiaolin Hu, and
  Jianguo Li.
\newblock Boosting adversarial attacks with momentum.
\newblock In \emph{Proceedings of the IEEE conference on computer vision and
  pattern recognition}, pp.\  9185--9193, 2018.

\bibitem[Engstrom et~al.(2018)Engstrom, Tran, Tsipras, Schmidt, and
  Madry]{engstrom2018rotation}
Logan Engstrom, Brandon Tran, Dimitris Tsipras, Ludwig Schmidt, and Aleksander
  Madry.
\newblock A rotation and a translation suffice: Fooling cnns with simple
  transformations.
\newblock 2018.

\bibitem[Ganin \& Lempitsky(2015)Ganin and Lempitsky]{ganin2015unsupervised}
Yaroslav Ganin and Victor Lempitsky.
\newblock Unsupervised domain adaptation by backpropagation.
\newblock In \emph{International conference on machine learning}, pp.\
  1180--1189. PMLR, 2015.

\bibitem[Ganin et~al.(2016)Ganin, Ustinova, Ajakan, Germain, Larochelle,
  Laviolette, Marchand, and Lempitsky]{ganin2016domain}
Yaroslav Ganin, Evgeniya Ustinova, Hana Ajakan, Pascal Germain, Hugo
  Larochelle, Fran{\c{c}}ois Laviolette, Mario Marchand, and Victor Lempitsky.
\newblock Domain-adversarial training of neural networks.
\newblock \emph{The journal of machine learning research}, 17\penalty0
  (1):\penalty0 2096--2030, 2016.

\bibitem[Gilmer et~al.(2018)Gilmer, Adams, Goodfellow, Andersen, and
  Dahl]{gilmer2018motivating}
Justin Gilmer, Ryan~P Adams, Ian Goodfellow, David Andersen, and George~E Dahl.
\newblock Motivating the rules of the game for adversarial example research.
\newblock \emph{arXiv preprint arXiv:1807.06732}, 2018.

\bibitem[Goldblum et~al.(2020)Goldblum, Fowl, Feizi, and
  Goldstein]{goldblum2020adversarially}
Micah Goldblum, Liam Fowl, Soheil Feizi, and Tom Goldstein.
\newblock Adversarially robust distillation.
\newblock In \emph{Proceedings of the AAAI Conference on Artificial
  Intelligence}, volume~34, pp.\  3996--4003, 2020.

\bibitem[Goodfellow et~al.(2014)Goodfellow, Shlens, and
  Szegedy]{goodfellow2014explaining}
Ian~J Goodfellow, Jonathon Shlens, and Christian Szegedy.
\newblock Explaining and harnessing adversarial examples.
\newblock \emph{arXiv preprint arXiv:1412.6572}, 2014.

\bibitem[Gowal et~al.(2018)Gowal, Dvijotham, Stanforth, Bunel, Qin, Uesato,
  Arandjelovic, Mann, and Kohli]{gowal2018effectiveness}
Sven Gowal, Krishnamurthy Dvijotham, Robert Stanforth, Rudy Bunel, Chongli Qin,
  Jonathan Uesato, Relja Arandjelovic, Timothy Mann, and Pushmeet Kohli.
\newblock On the effectiveness of interval bound propagation for training
  verifiably robust models.
\newblock \emph{arXiv preprint arXiv:1810.12715}, 2018.

\bibitem[Guo et~al.(2020)Guo, Yang, Xu, Liu, and Lin]{guo2020meets}
Minghao Guo, Yuzhe Yang, Rui Xu, Ziwei Liu, and Dahua Lin.
\newblock When nas meets robustness: In search of robust architectures against
  adversarial attacks.
\newblock In \emph{Proceedings of the IEEE/CVF Conference on Computer Vision
  and Pattern Recognition}, pp.\  631--640, 2020.

\bibitem[He et~al.(2016)He, Zhang, Ren, and Sun]{he2016identity}
Kaiming He, Xiangyu Zhang, Shaoqing Ren, and Jian Sun.
\newblock Identity mappings in deep residual networks.
\newblock In \emph{European conference on computer vision}, pp.\  630--645.
  Springer, 2016.

\bibitem[Hendrycks \& Dietterich(2018)Hendrycks and
  Dietterich]{hendrycks2018benchmarking}
Dan Hendrycks and Thomas~G Dietterich.
\newblock Benchmarking neural network robustness to common corruptions and
  surface variations.
\newblock \emph{arXiv preprint arXiv:1807.01697}, 2018.

\bibitem[Jiang et~al.(2020)Jiang, Chen, Chen, and Wang]{jiang2020robust}
Ziyu Jiang, Tianlong Chen, Ting Chen, and Zhangyang Wang.
\newblock Robust pre-training by adversarial contrastive learning.
\newblock In \emph{NeurIPS}, 2020.

\bibitem[Jin et~al.(2022)Jin, Yi, Huang, Schewe, and Huang]{jin2022enhancing}
Gaojie Jin, Xinping Yi, Wei Huang, Sven Schewe, and Xiaowei Huang.
\newblock Enhancing adversarial training with second-order statistics of
  weights.
\newblock In \emph{Proceedings of the IEEE/CVF Conference on Computer Vision
  and Pattern Recognition}, pp.\  15273--15283, 2022.

\bibitem[Kannan et~al.(2018)Kannan, Kurakin, and
  Goodfellow]{kannan2018adversarial}
Harini Kannan, Alexey Kurakin, and Ian Goodfellow.
\newblock Adversarial logit pairing.
\newblock \emph{arXiv preprint arXiv:1803.06373}, 2018.

\bibitem[Khim \& Loh(2018)Khim and Loh]{khim2018adversarial}
Justin Khim and Po-Ling Loh.
\newblock Adversarial risk bounds for binary classification via function
  transformation.
\newblock \emph{arXiv preprint arXiv:1810.09519}, 2, 2018.

\bibitem[Krizhevsky et~al.(2009)Krizhevsky, Hinton,
  et~al.]{krizhevsky2009learning}
Alex Krizhevsky, Geoffrey Hinton, et~al.
\newblock Learning multiple layers of features from tiny images.
\newblock 2009.

\bibitem[Kurakin et~al.(2016{\natexlab{a}})Kurakin, Goodfellow, and
  Bengio]{kurakin2016atscale}
Alexey Kurakin, Ian Goodfellow, and Samy Bengio.
\newblock Adversarial machine learning at scale.
\newblock \emph{arXiv preprint arXiv:1611.01236}, 2016{\natexlab{a}}.

\bibitem[Kurakin et~al.(2016{\natexlab{b}})Kurakin, Goodfellow, Bengio,
  et~al.]{kurakin2016adversarial}
Alexey Kurakin, Ian Goodfellow, Samy Bengio, et~al.
\newblock Adversarial examples in the physical world, 2016{\natexlab{b}}.

\bibitem[LeCun et~al.(1998)LeCun, Bottou, Bengio, and
  Haffner]{lecun1998gradient}
Yann LeCun, L{\'e}on Bottou, Yoshua Bengio, and Patrick Haffner.
\newblock Gradient-based learning applied to document recognition.
\newblock \emph{Proceedings of the IEEE}, 86\penalty0 (11):\penalty0
  2278--2324, 1998.

\bibitem[Lee et~al.(2020)Lee, Lee, and Yoon]{lee2020adversarial}
Saehyung Lee, Hyungyu Lee, and Sungroh Yoon.
\newblock Adversarial vertex mixup: Toward better adversarially robust
  generalization.
\newblock In \emph{Proceedings of the IEEE/CVF Conference on Computer Vision
  and Pattern Recognition}, pp.\  272--281, 2020.

\bibitem[Lee et~al.(2021)Lee, Park, Lee, Yi, Lee, and Yoon]{lee2021removing}
Saehyung Lee, Changhwa Park, Hyungyu Lee, Jihun Yi, Jonghyun Lee, and Sungroh
  Yoon.
\newblock Removing undesirable feature contributions using out-of-distribution
  data.
\newblock \emph{arXiv preprint arXiv:2101.06639}, 2021.

\bibitem[Liu et~al.(2020)Liu, Tang, Liu, Chen, Huang, Tu, Song, and
  Tao]{liu2020towards}
Aishan Liu, Shiyu Tang, Xianglong Liu, Xinyun Chen, Lei Huang, Zhuozhuo Tu,
  Dawn Song, and Dacheng Tao.
\newblock Towards defending multiple adversarial perturbations via gated batch
  normalization.
\newblock \emph{arXiv preprint arXiv:2012.01654}, 2020.

\bibitem[Madry et~al.(2017)Madry, Makelov, Schmidt, Tsipras, and
  Vladu]{madry2017towards}
Aleksander Madry, Aleksandar Makelov, Ludwig Schmidt, Dimitris Tsipras, and
  Adrian Vladu.
\newblock Towards deep learning models resistant to adversarial attacks.
\newblock \emph{arXiv preprint arXiv:1706.06083}, 2017.

\bibitem[Mansour et~al.(2009)Mansour, Mohri, and
  Rostamizadeh]{mansour2009domain}
Yishay Mansour, Mehryar Mohri, and Afshin Rostamizadeh.
\newblock Domain adaptation: Learning bounds and algorithms.
\newblock \emph{arXiv preprint arXiv:0902.3430}, 2009.

\bibitem[Montasser et~al.(2019)Montasser, Hanneke, and Srebro]{montasser2019vc}
Omar Montasser, Steve Hanneke, and Nathan Srebro.
\newblock Vc classes are adversarially robustly learnable, but only improperly.
\newblock In \emph{Conference on Learning Theory}, pp.\  2512--2530, 2019.

\bibitem[Moosavi-Dezfooli et~al.(2016)Moosavi-Dezfooli, Fawzi, and
  Frossard]{moosavi2016deepfool}
Seyed-Mohsen Moosavi-Dezfooli, Alhussein Fawzi, and Pascal Frossard.
\newblock Deepfool: a simple and accurate method to fool deep neural networks.
\newblock In \emph{Proceedings of the IEEE conference on computer vision and
  pattern recognition}, pp.\  2574--2582, 2016.

\bibitem[Netzer et~al.(2011)Netzer, Wang, Coates, Bissacco, Wu, and
  Ng]{netzer2011reading}
Yuval Netzer, Tao Wang, Adam Coates, Alessandro Bissacco, Bo~Wu, and Andrew~Y
  Ng.
\newblock Reading digits in natural images with unsupervised feature learning.
\newblock 2011.

\bibitem[Pang et~al.(2019)Pang, Xu, Du, Chen, and Zhu]{pang2019improving}
Tianyu Pang, Kun Xu, Chao Du, Ning Chen, and Jun Zhu.
\newblock Improving adversarial robustness via promoting ensemble diversity.
\newblock In \emph{International Conference on Machine Learning}, pp.\
  4970--4979. PMLR, 2019.

\bibitem[Pang et~al.(2020{\natexlab{a}})Pang, Yang, Dong, Su, and
  Zhu]{pang2020bag}
Tianyu Pang, Xiao Yang, Yinpeng Dong, Hang Su, and Jun Zhu.
\newblock Bag of tricks for adversarial training.
\newblock \emph{arXiv preprint arXiv:2010.00467}, 2020{\natexlab{a}}.

\bibitem[Pang et~al.(2020{\natexlab{b}})Pang, Yang, Dong, Xu, Zhu, and
  Su]{pang2020boosting}
Tianyu Pang, Xiao Yang, Yinpeng Dong, Kun Xu, Jun Zhu, and Hang Su.
\newblock Boosting adversarial training with hypersphere embedding.
\newblock \emph{arXiv preprint arXiv:2002.08619}, 2020{\natexlab{b}}.

\bibitem[Prabhu et~al.(2019)Prabhu, Yap, Xu, and
  Whaley]{prabhu2019understanding}
Vinay~Uday Prabhu, Dian~Ang Yap, Joyce Xu, and John Whaley.
\newblock Understanding adversarial robustness through loss landscape
  geometries.
\newblock \emph{arXiv preprint arXiv:1907.09061}, 2019.

\bibitem[Qian et~al.(2021)Qian, Zhang, Huang, Wang, Zhang, and
  Yi]{qian2021improving}
Zhuang Qian, Shufei Zhang, Kaizhu Huang, Qiufeng Wang, Rui Zhang, and Xinping
  Yi.
\newblock Improving model robustness with latent distribution locally and
  globally.
\newblock \emph{arXiv preprint arXiv:2107.04401}, 2021.

\bibitem[Raghunathan et~al.(2018{\natexlab{a}})Raghunathan, Steinhardt, and
  Liang]{raghunathan2018certified}
Aditi Raghunathan, Jacob Steinhardt, and Percy Liang.
\newblock Certified defenses against adversarial examples.
\newblock \emph{arXiv preprint arXiv:1801.09344}, 2018{\natexlab{a}}.

\bibitem[Raghunathan et~al.(2018{\natexlab{b}})Raghunathan, Steinhardt, and
  Liang]{raghunathan2018semidefinite}
Aditi Raghunathan, Jacob Steinhardt, and Percy Liang.
\newblock Semidefinite relaxations for certifying robustness to adversarial
  examples.
\newblock \emph{arXiv preprint arXiv:1811.01057}, 2018{\natexlab{b}}.

\bibitem[Rauber et~al.(2017)Rauber, Brendel, and Bethge]{rauber2017foolbox}
Jonas Rauber, Wieland Brendel, and Matthias Bethge.
\newblock Foolbox: A python toolbox to benchmark the robustness of machine
  learning models.
\newblock In \emph{Reliable Machine Learning in the Wild Workshop, 34th
  International Conference on Machine Learning}, 2017.
\newblock URL \url{http://arxiv.org/abs/1707.04131}.

\bibitem[Rebuffi et~al.(2021{\natexlab{a}})Rebuffi, Gowal, Calian, Stimberg,
  Wiles, and Mann]{rebuffi2021fixing}
Sylvestre-Alvise Rebuffi, Sven Gowal, Dan~A Calian, Florian Stimberg, Olivia
  Wiles, and Timothy Mann.
\newblock Fixing data augmentation to improve adversarial robustness.
\newblock \emph{arXiv preprint arXiv:2103.01946}, 2021{\natexlab{a}}.

\bibitem[Rebuffi et~al.(2021{\natexlab{b}})Rebuffi, Gowal, Calian, Stimberg,
  Wiles, and Mann]{rebuffi2021data}
Sylvestre-Alvise Rebuffi, Sven Gowal, Dan~Andrei Calian, Florian Stimberg,
  Olivia Wiles, and Timothy~A Mann.
\newblock Data augmentation can improve robustness.
\newblock \emph{Advances in Neural Information Processing Systems}, 34,
  2021{\natexlab{b}}.

\bibitem[Rice et~al.(2020)Rice, Wong, and Kolter]{rice2020overfitting}
Leslie Rice, Eric Wong, and Zico Kolter.
\newblock Overfitting in adversarially robust deep learning.
\newblock In \emph{International Conference on Machine Learning}, pp.\
  8093--8104. PMLR, 2020.

\bibitem[Rony et~al.(2019)Rony, Hafemann, Oliveira, Ayed, Sabourin, and
  Granger]{rony2019decoupling}
J{\'e}r{\^o}me Rony, Luiz~G Hafemann, Luiz~S Oliveira, Ismail~Ben Ayed, Robert
  Sabourin, and Eric Granger.
\newblock Decoupling direction and norm for efficient gradient-based l2
  adversarial attacks and defenses.
\newblock In \emph{Proceedings of the IEEE/CVF Conference on Computer Vision
  and Pattern Recognition}, pp.\  4322--4330, 2019.

\bibitem[Salman et~al.(2020)Salman, Ilyas, Engstrom, Kapoor, and
  Madry]{salman2020adversarially}
Hadi Salman, Andrew Ilyas, Logan Engstrom, Ashish Kapoor, and Aleksander Madry.
\newblock Do adversarially robust imagenet models transfer better?
\newblock \emph{arXiv preprint arXiv:2007.08489}, 2020.

\bibitem[Schmidt et~al.(2018)Schmidt, Santurkar, Tsipras, Talwar, and
  M{\k{a}}dry]{schmidt2018adversarially}
Ludwig Schmidt, Shibani Santurkar, Dimitris Tsipras, Kunal Talwar, and
  Aleksander M{\k{a}}dry.
\newblock Adversarially robust generalization requires more data.
\newblock \emph{arXiv preprint arXiv:1804.11285}, 2018.

\bibitem[Shafahi et~al.(2019)Shafahi, Najibi, Ghiasi, Xu, Dickerson, Studer,
  Davis, Taylor, and Goldstein]{shafahi2019adversarial}
Ali Shafahi, Mahyar Najibi, Amin Ghiasi, Zheng Xu, John Dickerson, Christoph
  Studer, Larry~S Davis, Gavin Taylor, and Tom Goldstein.
\newblock Adversarial training for free!
\newblock \emph{arXiv preprint arXiv:1904.12843}, 2019.

\bibitem[Sinha et~al.(2017)Sinha, Namkoong, and Duchi]{sinha2017certifiable}
Aman Sinha, Hongseok Namkoong, and John Duchi.
\newblock Certifiable distributional robustness with principled adversarial
  training.
\newblock \emph{arXiv preprint arXiv:1710.10571}, 2, 2017.

\bibitem[Song et~al.(2018)Song, He, Wang, and Hopcroft]{song2018improving}
Chuanbiao Song, Kun He, Liwei Wang, and John~E Hopcroft.
\newblock Improving the generalization of adversarial training with domain
  adaptation.
\newblock \emph{arXiv preprint arXiv:1810.00740}, 2018.

\bibitem[Sun \& Saenko(2016)Sun and Saenko]{sun2016deep}
Baochen Sun and Kate Saenko.
\newblock Deep coral: Correlation alignment for deep domain adaptation.
\newblock In \emph{European conference on computer vision}, pp.\  443--450.
  Springer, 2016.

\bibitem[Szegedy et~al.(2013)Szegedy, Zaremba, Sutskever, Bruna, Erhan,
  Goodfellow, and Fergus]{szegedy2013intriguing}
Christian Szegedy, Wojciech Zaremba, Ilya Sutskever, Joan Bruna, Dumitru Erhan,
  Ian Goodfellow, and Rob Fergus.
\newblock Intriguing properties of neural networks.
\newblock \emph{arXiv preprint arXiv:1312.6199}, 2013.

\bibitem[Tabacof \& Valle(2016)Tabacof and Valle]{tabacof2016exploring}
Pedro Tabacof and Eduardo Valle.
\newblock Exploring the space of adversarial images.
\newblock In \emph{2016 International Joint Conference on Neural Networks
  (IJCNN)}, pp.\  426--433. IEEE, 2016.

\bibitem[Tram{\`e}r et~al.(2017)Tram{\`e}r, Kurakin, Papernot, Goodfellow,
  Boneh, and McDaniel]{tramer2017ensemble}
Florian Tram{\`e}r, Alexey Kurakin, Nicolas Papernot, Ian Goodfellow, Dan
  Boneh, and Patrick McDaniel.
\newblock Ensemble adversarial training: Attacks and defenses.
\newblock \emph{arXiv preprint arXiv:1705.07204}, 2017.

\bibitem[Tsai et~al.(2021)Tsai, Hsu, Yu, and Chen]{tsai2021formalizing}
Yu-Lin Tsai, Chia-Yi Hsu, Chia-Mu Yu, and Pin-Yu Chen.
\newblock Formalizing generalization and robustness of neural networks to
  weight perturbations.
\newblock \emph{arXiv preprint arXiv:2103.02200}, 2021.

\bibitem[Tsipras et~al.(2018)Tsipras, Santurkar, Engstrom, Turner, and
  Madry]{tsipras2018robustness}
Dimitris Tsipras, Shibani Santurkar, Logan Engstrom, Alexander Turner, and
  Aleksander Madry.
\newblock Robustness may be at odds with accuracy.
\newblock \emph{arXiv preprint arXiv:1805.12152}, 2018.

\bibitem[Uesato et~al.(2019)Uesato, Alayrac, Huang, Stanforth, Fawzi, and
  Kohli]{uesato2019labels}
Jonathan Uesato, Jean-Baptiste Alayrac, Po-Sen Huang, Robert Stanforth,
  Alhussein Fawzi, and Pushmeet Kohli.
\newblock Are labels required for improving adversarial robustness?
\newblock \emph{arXiv preprint arXiv:1905.13725}, 2019.

\bibitem[Utrera et~al.(2020)Utrera, Kravitz, Erichson, Khanna, and
  Mahoney]{utrera2020adversarially}
Francisco Utrera, Evan Kravitz, N~Benjamin Erichson, Rajiv Khanna, and
  Michael~W Mahoney.
\newblock Adversarially-trained deep nets transfer better.
\newblock \emph{arXiv preprint arXiv:2007.05869}, 2020.

\bibitem[Van~der Maaten \& Hinton(2008)Van~der Maaten and
  Hinton]{van2008visualizing}
Laurens Van~der Maaten and Geoffrey Hinton.
\newblock Visualizing data using t-sne.
\newblock \emph{Journal of machine learning research}, 9\penalty0 (11), 2008.

\bibitem[Wang et~al.(2021)Wang, Geng, Jiang, Li, Wang, Yang, and
  Lin]{wang2021residual}
Yifei Wang, Zhengyang Geng, Feng Jiang, Chuming Li, Yisen Wang, Jiansheng Yang,
  and Zhouchen Lin.
\newblock Residual relaxation for multi-view representation learning.
\newblock \emph{Advances in Neural Information Processing Systems},
  34:\penalty0 12104--12115, 2021.

\bibitem[Wang et~al.(2019{\natexlab{a}})Wang, Ma, Bailey, Yi, Zhou, and
  Gu]{wang2019convergence}
Yisen Wang, Xingjun Ma, James Bailey, Jinfeng Yi, Bowen Zhou, and Quanquan Gu.
\newblock On the convergence and robustness of adversarial training.
\newblock In \emph{ICML}, volume~1, pp.\ ~2, 2019{\natexlab{a}}.

\bibitem[Wang et~al.(2019{\natexlab{b}})Wang, Zou, Yi, Bailey, Ma, and
  Gu]{wang2019improving}
Yisen Wang, Difan Zou, Jinfeng Yi, James Bailey, Xingjun Ma, and Quanquan Gu.
\newblock Improving adversarial robustness requires revisiting misclassified
  examples.
\newblock In \emph{International Conference on Learning Representations},
  2019{\natexlab{b}}.

\bibitem[Wong \& Kolter(2018)Wong and Kolter]{wong2018provable}
Eric Wong and Zico Kolter.
\newblock Provable defenses against adversarial examples via the convex outer
  adversarial polytope.
\newblock In \emph{International Conference on Machine Learning}, pp.\
  5286--5295. PMLR, 2018.

\bibitem[Wong et~al.(2018)Wong, Schmidt, Metzen, and Kolter]{wong2018scaling}
Eric Wong, Frank~R Schmidt, Jan~Hendrik Metzen, and J~Zico Kolter.
\newblock Scaling provable adversarial defenses.
\newblock \emph{arXiv preprint arXiv:1805.12514}, 2018.

\bibitem[Wong et~al.(2020)Wong, Rice, and Kolter]{wong2020fast}
Eric Wong, Leslie Rice, and J~Zico Kolter.
\newblock Fast is better than free: Revisiting adversarial training.
\newblock \emph{arXiv preprint arXiv:2001.03994}, 2020.

\bibitem[Wu et~al.(2020)Wu, Xia, and Wang]{wu2020adversarial}
Dongxian Wu, Shu-Tao Xia, and Yisen Wang.
\newblock Adversarial weight perturbation helps robust generalization.
\newblock \emph{Advances in Neural Information Processing Systems}, 33, 2020.

\bibitem[Xiao et~al.(2018)Xiao, Zhu, Li, He, Liu, and Song]{xiao2018spatially}
Chaowei Xiao, Jun-Yan Zhu, Bo~Li, Warren He, Mingyan Liu, and Dawn Song.
\newblock Spatially transformed adversarial examples.
\newblock \emph{arXiv preprint arXiv:1801.02612}, 2018.

\bibitem[Xie \& Yuille(2019)Xie and Yuille]{xie2019intriguing}
Cihang Xie and Alan Yuille.
\newblock Intriguing properties of adversarial training at scale.
\newblock \emph{arXiv preprint arXiv:1906.03787}, 2019.

\bibitem[Xie et~al.(2019{\natexlab{a}})Xie, Wu, Maaten, Yuille, and
  He]{xie2019feature}
Cihang Xie, Yuxin Wu, Laurens van~der Maaten, Alan~L Yuille, and Kaiming He.
\newblock Feature denoising for improving adversarial robustness.
\newblock In \emph{Proceedings of the IEEE/CVF Conference on Computer Vision
  and Pattern Recognition}, pp.\  501--509, 2019{\natexlab{a}}.

\bibitem[Xie et~al.(2019{\natexlab{b}})Xie, Zhang, Zhou, Bai, Wang, Ren, and
  Yuille]{xie2019improving}
Cihang Xie, Zhishuai Zhang, Yuyin Zhou, Song Bai, Jianyu Wang, Zhou Ren, and
  Alan~L Yuille.
\newblock Improving transferability of adversarial examples with input
  diversity.
\newblock In \emph{Proceedings of the IEEE/CVF Conference on Computer Vision
  and Pattern Recognition}, pp.\  2730--2739, 2019{\natexlab{b}}.

\bibitem[Yang et~al.(2020)Yang, Zhang, Dong, Inkawhich, Gardner, Touchet,
  Wilkes, Berry, and Li]{yang2020dverge}
Huanrui Yang, Jingyang Zhang, Hongliang Dong, Nathan Inkawhich, Andrew Gardner,
  Andrew Touchet, Wesley Wilkes, Heath Berry, and Hai Li.
\newblock Dverge: diversifying vulnerabilities for enhanced robust generation
  of ensembles.
\newblock \emph{arXiv preprint arXiv:2009.14720}, 2020.

\bibitem[Yin et~al.(2019)Yin, Kannan, and Bartlett]{yin2019rademacher}
Dong Yin, Ramchandran Kannan, and Peter Bartlett.
\newblock Rademacher complexity for adversarially robust generalization.
\newblock In \emph{International Conference on Machine Learning}, pp.\
  7085--7094. PMLR, 2019.

\bibitem[Yu et~al.(2018)Yu, Liu, Wang, Zhao, and Chen]{yu2018interpreting}
Fuxun Yu, Chenchen Liu, Yanzhi Wang, Liang Zhao, and Xiang Chen.
\newblock Interpreting adversarial robustness: A view from decision surface in
  input space.
\newblock \emph{arXiv preprint arXiv:1810.00144}, 2018.

\bibitem[Zagoruyko \& Komodakis(2016)Zagoruyko and
  Komodakis]{zagoruyko2016wide}
Sergey Zagoruyko and Nikos Komodakis.
\newblock Wide residual networks.
\newblock \emph{arXiv preprint arXiv:1605.07146}, 2016.

\bibitem[Zhai et~al.(2019)Zhai, Cai, He, Dan, He, Hopcroft, and
  Wang]{zhai2019adversarially}
Runtian Zhai, Tianle Cai, Di~He, Chen Dan, Kun He, John Hopcroft, and Liwei
  Wang.
\newblock Adversarially robust generalization just requires more unlabeled
  data.
\newblock \emph{arXiv preprint arXiv:1906.00555}, 2019.

\bibitem[Zhang et~al.(2019{\natexlab{a}})Zhang, Zhang, Lu, Zhu, and
  Dong]{zhang2019you}
Dinghuai Zhang, Tianyuan Zhang, Yiping Lu, Zhanxing Zhu, and Bin Dong.
\newblock You only propagate once: Accelerating adversarial training via
  maximal principle.
\newblock \emph{arXiv preprint arXiv:1905.00877}, 2019{\natexlab{a}}.

\bibitem[Zhang \& Wang(2019)Zhang and Wang]{zhang2019defense}
Haichao Zhang and Jianyu Wang.
\newblock Defense against adversarial attacks using feature scattering-based
  adversarial training.
\newblock \emph{Advances in Neural Information Processing Systems},
  32:\penalty0 1831--1841, 2019.

\bibitem[Zhang et~al.(2019{\natexlab{b}})Zhang, Yu, Jiao, Xing, El~Ghaoui, and
  Jordan]{zhang2019theoretically}
Hongyang Zhang, Yaodong Yu, Jiantao Jiao, Eric Xing, Laurent El~Ghaoui, and
  Michael Jordan.
\newblock Theoretically principled trade-off between robustness and accuracy.
\newblock In \emph{International Conference on Machine Learning}, pp.\
  7472--7482. PMLR, 2019{\natexlab{b}}.

\bibitem[Zhang et~al.(2020)Zhang, Xu, Han, Niu, Cui, Sugiyama, and
  Kankanhalli]{zhang2020attacks}
Jingfeng Zhang, Xilie Xu, Bo~Han, Gang Niu, Lizhen Cui, Masashi Sugiyama, and
  Mohan Kankanhalli.
\newblock Attacks which do not kill training make adversarial learning
  stronger.
\newblock In \emph{International Conference on Machine Learning}, pp.\
  11278--11287. PMLR, 2020.

\end{thebibliography}
\bibliographystyle{tmlr}

\appendix
\newpage
\section{Domain Invariant Adversarial Learning Algorithm}
\label{algo-appendix}
Algorithm~\ref{DIAL-algorithm} describes a pseudo-code of our proposed $\DIAL_{\ce}$ variant. As can be seen, a target domain batch is not given in advance as with standard domain-adaptation task. Instead, for each natural batch we generate a target batch using adversarial training. The loss function is composed of natural and adversarial losses with respect to the main task (e.g., classification), and from natural and adversarial domain losses. By maximizing the losses on the domain we aim to learn a feature representation which is invariant to the natural and adversarial domain, and therefore more robust.

\begin{algorithm}[tb]

\begin{algorithmic}
 \STATE {\bfseries Input:} Source data $S={\{(x_i,y_i)\}}_{i=1}^n$ and network architecture $G_f,G_y,G_d$
 \STATE {\bfseries Parameters:} Batch size $m$, perturbation size $\epsilon$, pgd attack step size $\tau$, adversarial trade-off $\lambda$, initial reversal ratio $r$, and step size $\alpha$
 \STATE {\bfseries Init:} $Y_{0}$ and $Y_{1}$  source and target domain vectors filled with 0 and 1 respectively
 \STATE {\bfseries Output:} Robust network $G=(G_f,G_y,G_d)$ parameterized by $\hat{\theta}=(\theta_f,\theta_y, \theta_d)$ respectively

 \REPEAT 
  \STATE {Fetch mini-batch $X_{s}={\{x_j\}}_{j=1}^m$, $Y_{s}={\{y_j\}}_{j=1}^m$}
  
  \STATE{\# Generate adversarial target domain batch $X_{t}$}
  
  \FOR{$j=1,\ldots,m$ (in parallel)} \STATE{ 
    $x'_j \leftarrow PGD(x_j,y_j,\epsilon,\tau)$
   
    $X_{t} \leftarrow X_{t} + x'_j$
  }
  \ENDFOR

  \STATE{$\ell_{s}^y, \, \ell_{t}^y \leftarrow \text{CE}(G_y(G_f(X_{s})), Y_{s}), \, \text{CE}(G_y(G_f(X_{t})), Y_{s})$}

  \STATE{$\ell_{s}^d, \, \ell_{t}^d \leftarrow \text{CE}(G_d(G_f(X_{s})), Y_{0}), \, \text{CE}(G_d(G_f(X_{t})), Y_{1})$}
  
  \STATE{$\ell \leftarrow \ell_{s}^y + \lambda\ell_{t}^y - r(\ell_{s}^d + \ell_{t}^d)$}
  
  \STATE{$\hat{\theta} \leftarrow \hat{\theta} - \alpha\nabla_{\hat{\theta}}(\ell)$}
\UNTIL{stopping criterion is not met}

\caption{Domain Invariant Adversarial Learning}
\label{DIAL-algorithm}
\end{algorithmic}
\end{algorithm}

\section{additional results on CIFAR-100 and SVHN}
\label{cifar100-svhn-appendix}

\begin{table}[!ht]
  \caption{Robustness against white-box, black-box attacks and Auto-Attack (AA) on SVHN. Black-box attacks are generated using naturally trained surrogate model and applied to the best performing robust models.}
  \label{black-and_white-svhn-appendix}
  \vskip 0.15in
  \centering
  \small
  \begin{tabular}{l@{\hspace{1\tabcolsep}}c@{\hspace{1\tabcolsep}}c@{\hspace{1\tabcolsep}}c@{\hspace{1\tabcolsep}}c@{\hspace{1\tabcolsep}}c@{\hspace{1\tabcolsep}}c@{\hspace{1\tabcolsep}}c@{\hspace{1\tabcolsep}}c@{\hspace{1\tabcolsep}}c@{\hspace{1\tabcolsep}}c}
    \toprule
    & & \multicolumn{4}{c}{White-box} & \multicolumn{4}{c}{Black-Box}  \\
    \cmidrule(r){3-6} 
    \cmidrule(r){7-10}
    Defense Model & Natural & PGD$^{20}$ & PGD$^{100}$  & PGD$^{1000}$  & CW$^{\infty}$ & PGD$^{20}$ & PGD$^{100}$ & PGD$^{1000}$  & CW$^{\infty}$ & AA \\
    \midrule
    TRADES & 90.35 & 57.10 & 54.13 & 54.08 & 52.19 & 86.89 & 86.73 & 86.57 & 86.70 &  49.5 \\
    $\DIAL_{\kl}$ (Ours) & 90.66 & \textbf{58.91} & \textbf{55.30} & \textbf{55.11} & \textbf{53.67} & 87.62 & 87.52 & 87.41 & 87.63 & \textbf{51.00} \\
    $\DIAL_{\ce}$ (Ours) & \textbf{92.88} & 55.26  & 50.82 & 50.54 & 49.66 & \textbf{89.12} & \textbf{89.01} & \textbf{88.74} & \textbf{89.10} &  46.52  \\
    \bottomrule
  \end{tabular}
\end{table}

\begin{table}[!ht]
  \caption{Robustness against white-box, black-box attacks and Auto-Attack (AA) on CIFAR100. Black-box attacks are generated using naturally trained surrogate model and applied to the best performing robust models.}
  \label{black-and_white-cifar100-appendix}
  \vskip 0.15in
  \centering
  \small
  \begin{tabular}{l@{\hspace{1\tabcolsep}}c@{\hspace{1\tabcolsep}}c@{\hspace{1\tabcolsep}}c@{\hspace{1\tabcolsep}}c@{\hspace{1\tabcolsep}}c@{\hspace{1\tabcolsep}}c@{\hspace{1\tabcolsep}}c@{\hspace{1\tabcolsep}}c@{\hspace{1\tabcolsep}}c@{\hspace{1\tabcolsep}}c}
    \toprule
    & & \multicolumn{4}{c}{White-box} & \multicolumn{4}{c}{Black-Box}  \\
    \cmidrule(r){3-6} 
    \cmidrule(r){7-10}
    Defense Model & Natural & PGD$^{20}$ & PGD$^{100}$  & PGD$^{1000}$  & CW$^{\infty}$ & PGD$^{20}$ & PGD$^{100}$ & PGD$^{1000}$  & CW$^{\infty}$ & AA \\
    \midrule
    TRADES & 58.24 & 30.1 & 29.66 & 29.64 & 25.97 & 57.05 & 56.71 & 56.67 & 56.77 & 24.92 \\
    $\DIAL_{\kl}$ (Ours) & 58.47 & \textbf{31.19} & \textbf{30.50} & \textbf{30.42} & \textbf{26.91} & 57.16 & 56.81 & 56.80 & 57.00 & \textbf{25.87} \\
    $\DIAL_{\ce}$ (Ours) & \textbf{60.77} & 27.87 & 26.66 & 26.61 & 25.98 & \textbf{59.48} & \textbf{59.06} & \textbf{58.96} & \textbf{59.20} & 23.51  \\
    \bottomrule
  \end{tabular}
\end{table}

\newpage

\section{CIFAR-10 Additional Experimental Setup details}
\label{cifar10-additional-setup}
\paragraph{Additional defence setup.} For being consistent with other methods, the natural images are padded with 4-pixel padding with 32-random crop and random horizontal flip. Furthermore, all methods are trained using SGD with momentum 0.9. For $\DIAL_{\kl}$, we balance the robust loss with $\lambda=6$ and the domains loss with $r=4$. For $\DIAL_{\ce}$, we balance the robust loss with $\lambda=1$ and the domains loss with $r=2$. 
For DIAL-AWP, we used the same learning rate schedule used in ~\cite{wu2020adversarial}, where the initial 0.1 learning rate decays by a factor of 10 after 100 and 150 iterations. For black-box attacks, we used two types of surrogate models (1) naturally trained surrogate model, with natural accuracy of 95.61\% and (2) surrogate model trained using one of the adversarial training methods.

\section{Benchmarking the State-of-the-art on MNIST}
\label{mnist-results}
\paragraph{Defence setup.} We use the same CNN architecture as used in~\cite{zhang2019theoretically} which consists of four convolutional layers and three fully-connected layers. Sidelong this architecture, we integrate a domain classification layer. To generate the adversarial domain dataset, we use a perturbation size of $\epsilon=0.3$. We apply 40 iterations of inner maximization with perturbation step size of 0.01. Batch size is set to 128 and the model is trained for 100 epochs. Similar to the other methods, the initial learning rate was set to 0.01, and decays by a factor of 10 after 55 iterations, 75 and 90 iterations. All the models in the experiment are trained using SGD with momentum 0.9. For our method, we balance the robust loss with $\lambda=6$ and the domains loss with $r=0.1$. 

\paragraph{White-box/Black-box robustness.} We evaluate all defense models using PGD$^{40}$, PGD$^{100}$, PGD$^{1000}$ and CW$_{\infty}$ ($\ell_{\infty}$ version of~\cite{carlini2017towards} attack optimized by PGD-100) with step size 0.01. We constrain all attacks by the same perturbation $\epsilon=0.3$. 
For our black-box setting, we use a naturally trained surrogate model with natural accuracy of 99.51\%. As reported in Table~\ref{black-and_white-mnist}, our method achieves improved robustness over the other methods under the different attack types, while preserving the same level of natural accuracy, and even surpassing the naturally trained model. We should note that in general, the improvement margin on MNIST is more moderate compared to CIFAR-10, since MNIST is an easier task than CIFAR-10 and the robustness range is already high to begin with. Additional results are available in Appendix~\ref{additional_res}.

\begin{table}[!ht]
  \caption{Robustness against white-box, black-box attacks and Auto-Attack (AA) on MNIST. Black-box attacks are generated using naturally trained surrogate model and applied to the best performing robust models.}
  \label{black-and_white-mnist}
  \vskip 0.15in
  \centering
  \begin{tabular}{lllllllll}
    \toprule
    & & \multicolumn{3}{c}{White-box} & \multicolumn{3}{c}{Black-Box} \\
    \cmidrule(r){3-5} 
    \cmidrule(r){6-8}
    Defense Model & Natural & PGD$^{40}$ & PGD$^{100}$ & CW$^{\infty}$ & PGD$^{40}$ & PGD$^{100}$ & CW$^{\infty}$ & AA \\
    \midrule
    TRADES & 99.48 & 96.07 & 95.52 & 95.69 & 98.12 & 97.86 & 98.21 & 92.79 \\
    MART & 99.38 & 96.99 & 96.11 & 95.98 & 98.16 & 97.96 & 98.28 & 93.30 \\
    AT & 99.41 & 96.01 & 95.49 & 95.78 & 98.05 & 97.73 & 98.20 & 88.50 \\
    ATDA & 98.72 & 96.82 & 96.26 & 96.31 & 97.74 & 97.28 & 97.76 & 93.31 \\
    $\DIAL_{\kl}$ (Ours)  & 99.46 & 97.05 & 96.06 & 96.17 & 98.14 & 97.83 & 98.14 & \textbf{93.68} \\
    $\DIAL_{\ce}$ (Ours)  & $\mathbf{99.52}$ & $\mathbf{97.61}$ & $\mathbf{96.91}$ & $\mathbf{97.00}$ & $\mathbf{98.41}$ & $\mathbf{98.12}$ & $\mathbf{98.48}$ & 93.43 \\
    \bottomrule
  \end{tabular}
\end{table}

\section{Additional Results on MNIST and CIFAR-10} \label{additional_res}
In Table~\ref{pgd1000_attack} we present additional results using the PGD$^{1000}$ threat model. We use step size of 0.003 and constrain the attacks by the same perturbation $\epsilon=0.031$. Table~\ref{awp_variants} presents a comparison of our method combined with AWP to other the variants of AWP that were presented in ~\cite{wu2020adversarial}. In addition, in Table~\ref{f1-awp} we add the F$_{1}$-robust scores for different variants of AWP.


\begin{table}[!ht]
  \caption{PGD$^{1000}$ attack on MNIST and CIFAR-10 on white-box and black-box settings.}
  \label{pgd1000_attack}
  \vskip 0.15in
  \centering
  \begin{tabular}{llllll}
    \toprule
    & \multicolumn{2}{c}{MNIST} & \multicolumn{2}{c}{CIFAR-10} \\
    \cmidrule(r){2-3} 
    \cmidrule(r){4-5}
    Defense Model & White-box & Black-box & White-box & Black-box  \\
    \midrule
    TRADES & 95.22 & 97.81 & 56.43 & 83.80 \\
    MART & 95.74 & 97.89 & 56.55 & 82.47 \\
    AT & 95.36 & 97.78 & 54.40 & 83.96 \\
    ATDA & 96.20 & 97.34 & 41.02 & 75.11 \\
    $\DIAL_{\ce}$ (Ours) &  $\mathbf{96.78}$ & $\mathbf{98.10}$  & 51.57 & $\mathbf{88.22}$  \\
    $\DIAL_{\kl}$ (Ours) & 95.99 & 97.89 & $\mathbf{56.73}$ & 84.00 \\
    \bottomrule
  \end{tabular}
\end{table}

\begin{table}[!ht]
  \caption{Robustness comparison of DIAL-AWP and other variants of AWP that do not require additional data under the $\ell_{\infty}$ threat model.}
  \label{awp_variants}
  \vskip 0.15in
  \centering
  \begin{tabular}{lllll|l}
    \toprule
    \cmidrule(r){1-5}
    Defense Model & Natural & PGD$^{20}$ & PGD$^{100}$ & CW$_{\infty}$ & AA\\
    \midrule
    DIAL-AWP (Ours) & $\mathbf{85.91}$ & $\mathbf{61.10}$ & $\mathbf{59.86}$ & $\mathbf{57.67}$ & $\mathbf{56.78}$  \\
    TRADES-AWP \citep{wu2020adversarial} & 85.36 & 59.27 & 59.12 & 57.07 & 56.17    \\
    MART-AWP \citep{wu2020adversarial} & 84.43 & 60.68 & 59.32 & 56.37 & 54.23  \\
    AT-AWP \citep{wu2020adversarial} & 85.57  &  58.14 & 57.94 & 55.96 & 54.04 \\
    \bottomrule
  \end{tabular}
\end{table}

\begin{table}[!ht]
  \caption{F$_{1}$-robust measurement on AWP variants based on white-box attack.}
  \label{f1-awp}
  \vskip 0.15in
  \centering
  \begin{tabular}{ll}
    \toprule
    \cmidrule(r){1-2}
    Defense Model & F$_{1}$-robust \\
    \midrule
    DIAL-AWP (Ours) & $\mathbf{0.69753}$  \\
    TRADES-AWP \citep{wu2020adversarial} & 0.68162   \\
    MART-AWP \citep{wu2020adversarial} & 0.68857  \\
    AT-AWP \citep{wu2020adversarial} & 0.67381 \\
    \bottomrule
  \end{tabular}
\end{table}

\newpage
\section{Extened results on Unforeseen Corruptions}
\label{corruptions-apendix}
We present full accuracy results against unforeseen corruptions in Tables \ref{corruption-table1} and \ref{corruption-table2}. 

\begin{table}[!ht]
  \caption{Accuracy (\%) against unforeseen corruptions.}
  \label{corruption-table1}
  \vskip 0.15in
  \centering
  \tiny
  \begin{tabular}{lcccccccccccccccccc}
    \toprule
    Defense Model & brightness & defocus blur & fog & glass blur & jpeg compression & motion blur & saturate & snow & speckle noise  \\
    \midrule
    TRADES & 82.63 & 80.04 & 60.19 & 78.00 & 82.81 & 76.49 & 81.53 & 80.68 & 80.14 \\
    MART & 80.76 & 78.62 & 56.78 & 76.60 & 81.26 & 74.58 & 80.74 & 78.22 & 79.42 \\
    AT &  83.30 & 80.42 & 60.22 & 77.90 & 82.73 & 76.64 & 82.31 & 80.37 & 80.74 \\
    ATDA & 72.67 & 69.36 & 45.52 & 64.88 & 73.22 & 63.47 & 72.07 & 68.76 & 72.27 \\
    DIAL (Ours)  & \textbf{87.14} & \textbf{84.84} & \textbf{66.08} & \textbf{81.82} & \textbf{87.07} & \textbf{81.20} & \textbf{86.45} & \textbf{84.18} & \textbf{84.94} \\
    \bottomrule
  \end{tabular}
\end{table}

\begin{table}[!ht]
  \caption{Accuracy (\%) against unforeseen corruptions.}
  \label{corruption-table2}
  \vskip 0.15in
  \centering
  \tiny
  \begin{tabular}{lcccccccccccccccccc}
    \toprule
    Defense Model & contrast & elastic transform & frost & gaussian noise & impulse noise & pixelate & shot noise & spatter & zoom blur \\
    \midrule
    TRADES & 43.11 & 79.11 & 76.45 & 79.21 & 73.72 & 82.73 & 80.42 & 80.72 & 78.97 \\
    MART & 41.22 & 77.77 & 73.07 & 78.30 & 74.97 & 81.31 & 79.53 & 79.28 & 77.8 \\
    AT & 43.30 & 79.58 & 77.53 & 79.47 & 73.76 & 82.78 & 80.86 & 80.49 & 79.58 \\
    ATDA & 36.06 & 67.06 & 62.56 & 70.33 & 64.63 & 73.46 & 72.28 & 70.50 & 67.31 \\
    DIAL (Ours) & \textbf{48.84} & \textbf{84.13} & \textbf{81.76} & \textbf{83.76} & \textbf{78.26} & \textbf{87.24} & \textbf{85.13} & \textbf{84.84} & \textbf{83.93}  \\
    \bottomrule
  \end{tabular}
\end{table}


\section{Transfer Learning Settings}
\label{transfer-learning-settings}
The models used are the same models from previous experiments. We follow the common procedure of “fixed-feature” setting, where only a linear layer on top of the pre-trained network is trained. 
We train a linear classifier on CIFAR-100 on top of the pre-trained network which was trained on CIFAR-10. We also train a linear classifier on CIFAR-10 on top of the pre-trained nwork which was trained on CIFAR-100. We train the linear classifier for 100 epochs, and an initial learning rate of 0.1 which is decayed by a factor of 10 at epochs 50 and 75. We used SGD optimizer with momentum 0.9.

\section{Extended visualizations} \label{additional_viz}
\begin{figure}[H]
\centering
  \subfigure[\textbf{MART} embedded model output (logits) on natural test data]{\includegraphics[width=0.35\textwidth]{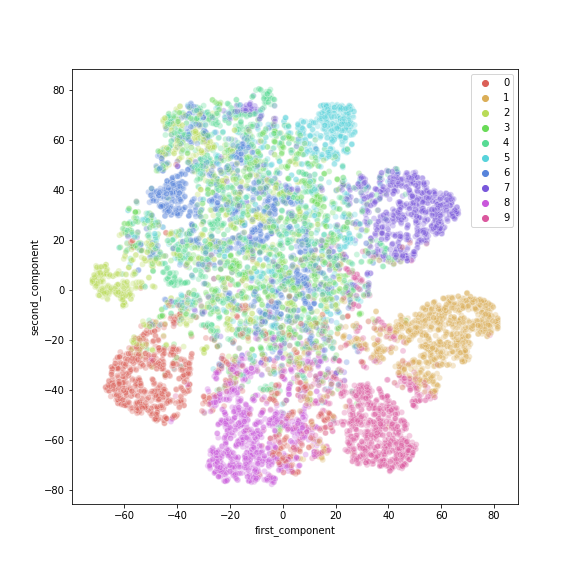}}
  \subfigure[\textbf{MART} embedded model output (logits) on adversarial test data]{\includegraphics[width=0.35\textwidth]{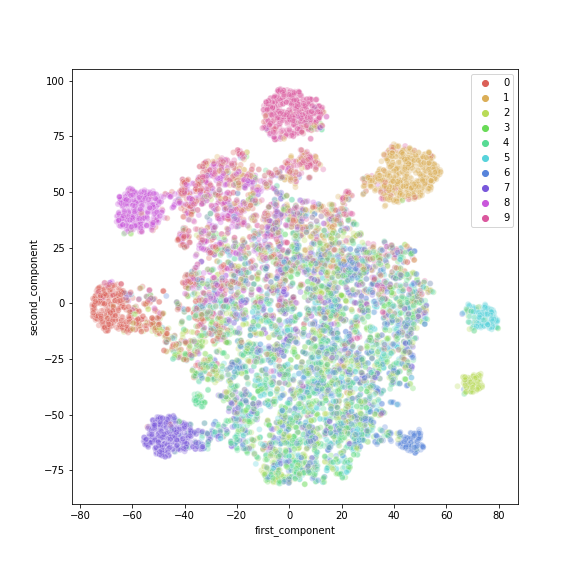}}
  
    \subfigure[\textbf{AT} embedded model output (logits) on natural test data]{\includegraphics[width=0.35\textwidth]{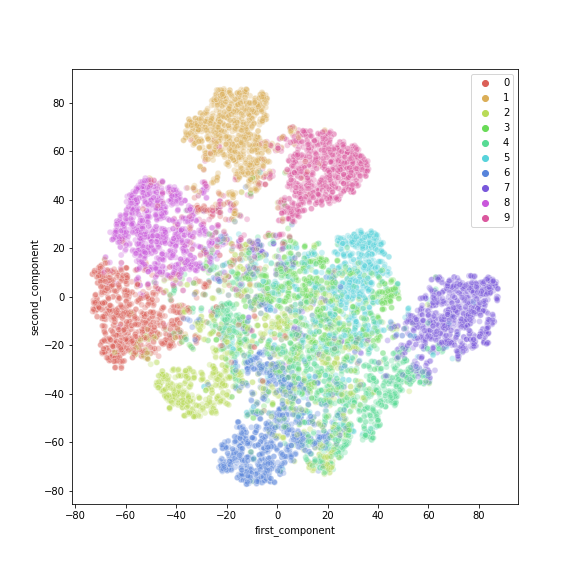}}
  \subfigure[\textbf{AT} embedded model output (logits) on adversarial test data]{\includegraphics[width=0.35\textwidth]{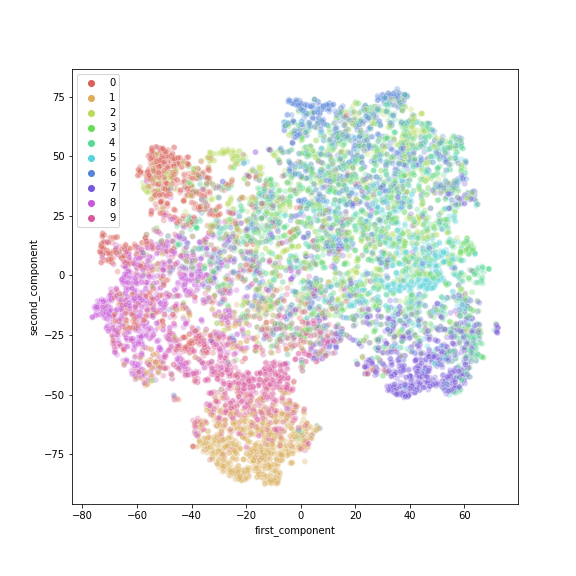}}
  
    \subfigure[\textbf{ATDA} embedded model output (logits) on natural test data]{\includegraphics[width=0.35\textwidth]{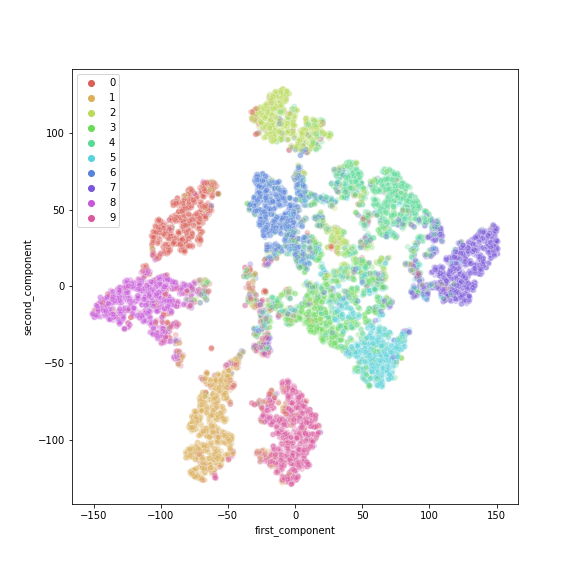}}
  \subfigure[\textbf{ATDA} embedded model output (logits) on adversarial test data]{\includegraphics[width=0.35\textwidth]{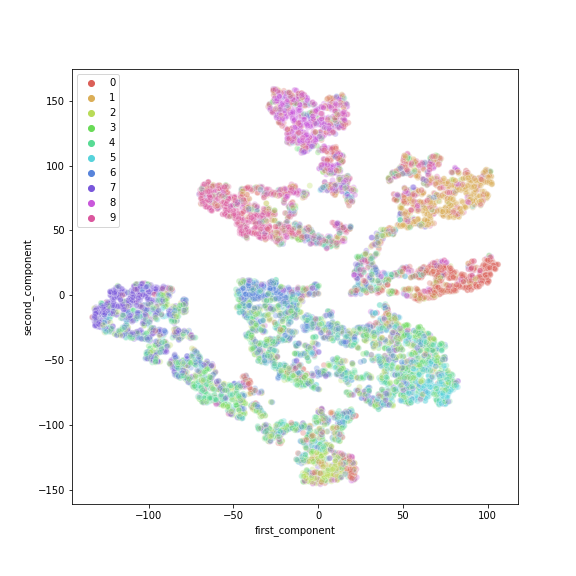}}
  \caption{t-SNE embedding of model output, \textbf{logits}, in two-dimensional space for MART, AT, and ATDA under natural and adversarial test data from CIFAR-10.}
  \label{tsne2}
\end{figure}

In Figure~\ref{tsne2}, we provide additional visualizations of the different adversarial training methods presented above. We visualize the models outputs (logits) using t-SNE with two components on the natural test data and the corresponding adversarial test data generated using PGD$^{20}$ white-box attack with step size 0.003 and $\epsilon=0.031$ on CIFAR-10. 

\begin{figure}[ht]
\centering
  \subfigure[Mean difference comparison]{\includegraphics[width=0.5\textwidth]{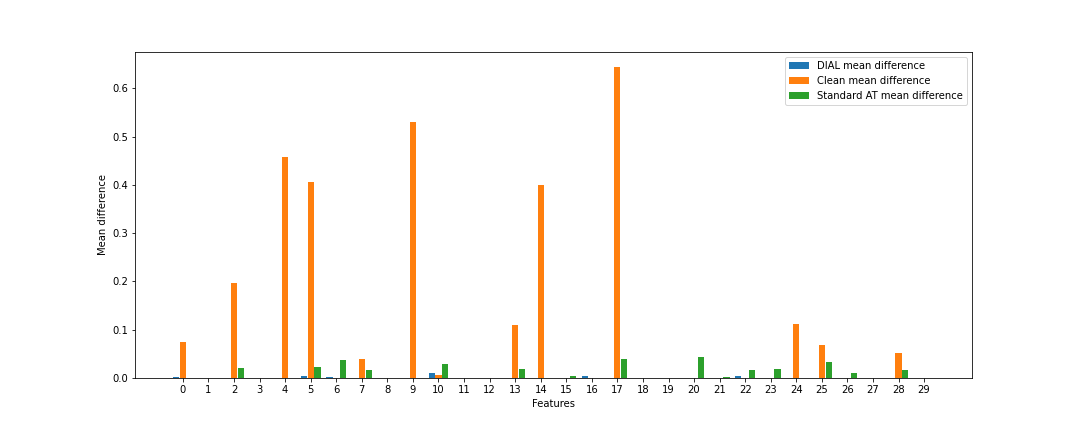}}
  \subfigure[Standard deviation difference comparison]{\includegraphics[width=0.5\textwidth]{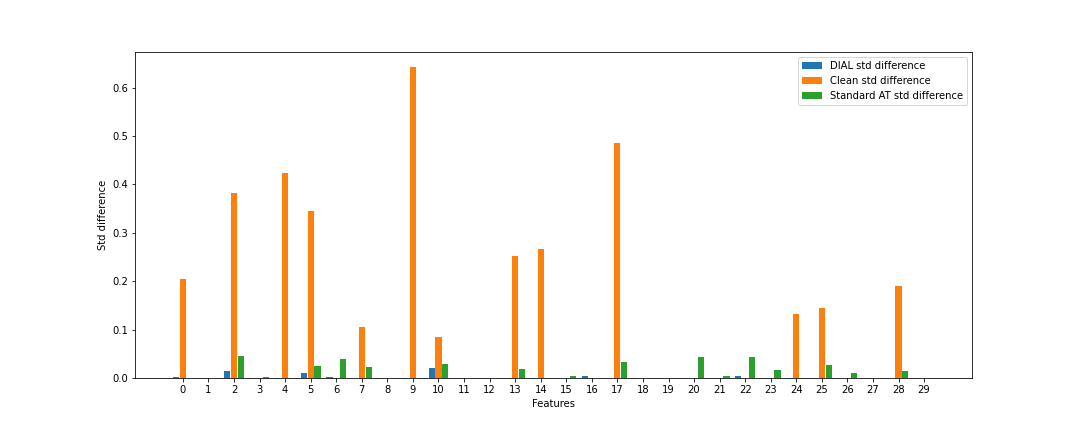}}
  \caption{Mean and std differences comparison between DIAL, naturally trained model and model trained using standard adversarial training on thirty random features from the features layer on the CIFAR-10 dataset with WRN-34-10 architecture. Each bar represents the absolute difference between the means/std of the natural examples and the mean/std of their corresponding adversarial examples on this same feature.}
  \label{feat_layer_stat2}
\end{figure}

In Figure \ref{feat_layer_stat2} we visualize statistical differences between natural and adversarial examples in the feature representation layer. Specifically, we show the differences in mean and std on thirty random feature values from the feature representation layer as we pass through a network the natural test examples and their corresponding adversarial examples. We present the results on same network architecture (WRN-34-10), trained using three different training procedures: naturally trained network, network trained using standard adversarial training (AT) \cite{madry2017towards}, and DIAL on the CIFAR-10 dataset. When the statistical characteristics of each feature differ from each other, it implies that the features layer is less domain invariant. That is, smaller differences in mean/std yields a better invariance to adversarial examples. One can observe that for DIAL, there is almost no differences between the mean/std of natural examples and their corresponding adversarial examples. Moreover, for the vast majority of the features, DIAL present smaller differences compared to the naturally trained model and the model trained with standard adversarial training. Best viewed in colors.


\newpage
\section{In-depth Analysis}
\begin{figure}[ht]
\centering
    \subfigure[\textbf{DIAL} embedded \textbf{features} on natural test data]{\includegraphics[width=0.35\textwidth]{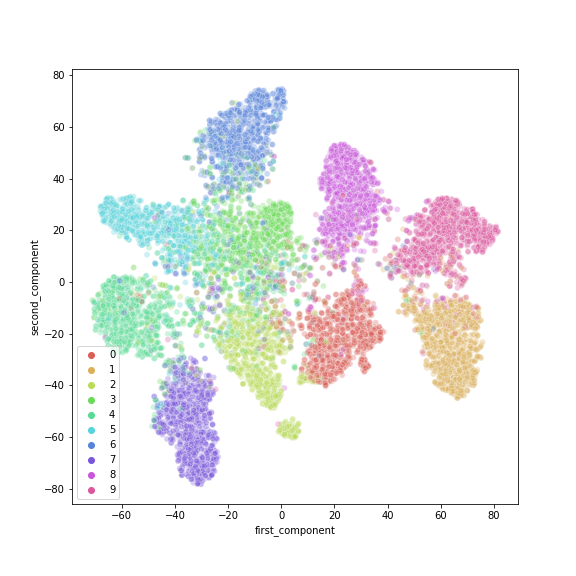}}
  \subfigure[\textbf{DIAL} embedded \textbf{features} on adversarial test data]{\includegraphics[width=0.35\textwidth]{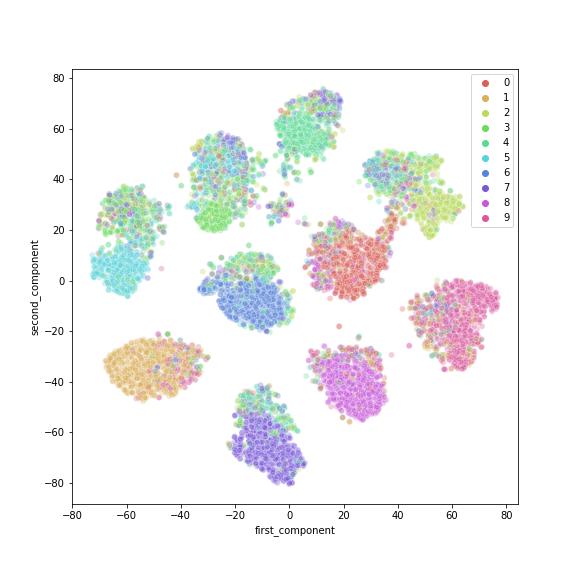}}

    \subfigure[\textbf{TRADES} embedded \textbf{features} on natural test data]{\includegraphics[width=0.35\textwidth]{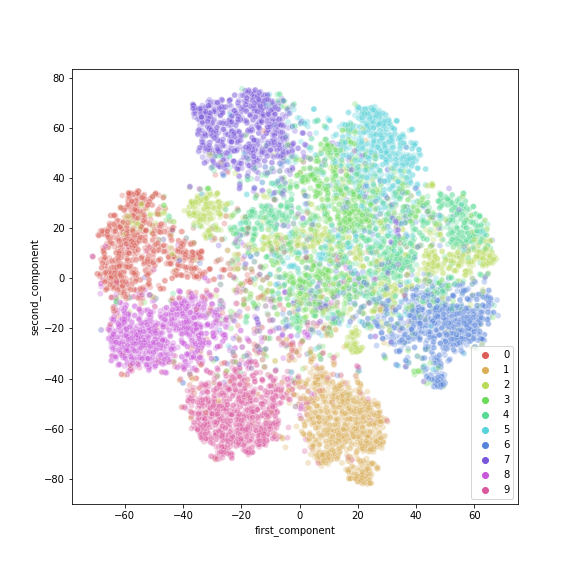}}
  \subfigure[\textbf{TRADES} embedded \textbf{features} on adversarial test data]{\includegraphics[width=0.35\textwidth]{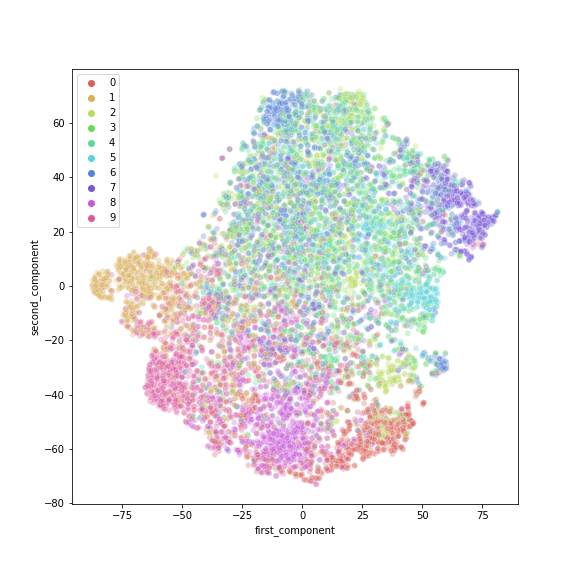}}

  \caption{Visualizing the two-dimensional T-SNE \textbf{feature space embedding} of DIAL and TRADES for (1) natural test data, and (2) adversarial test data from CIFAR-10.}
  \label{tsne-feat}
\end{figure}

\begin{figure}[!ht]
  \centering
  \includegraphics[width=0.8\textwidth]{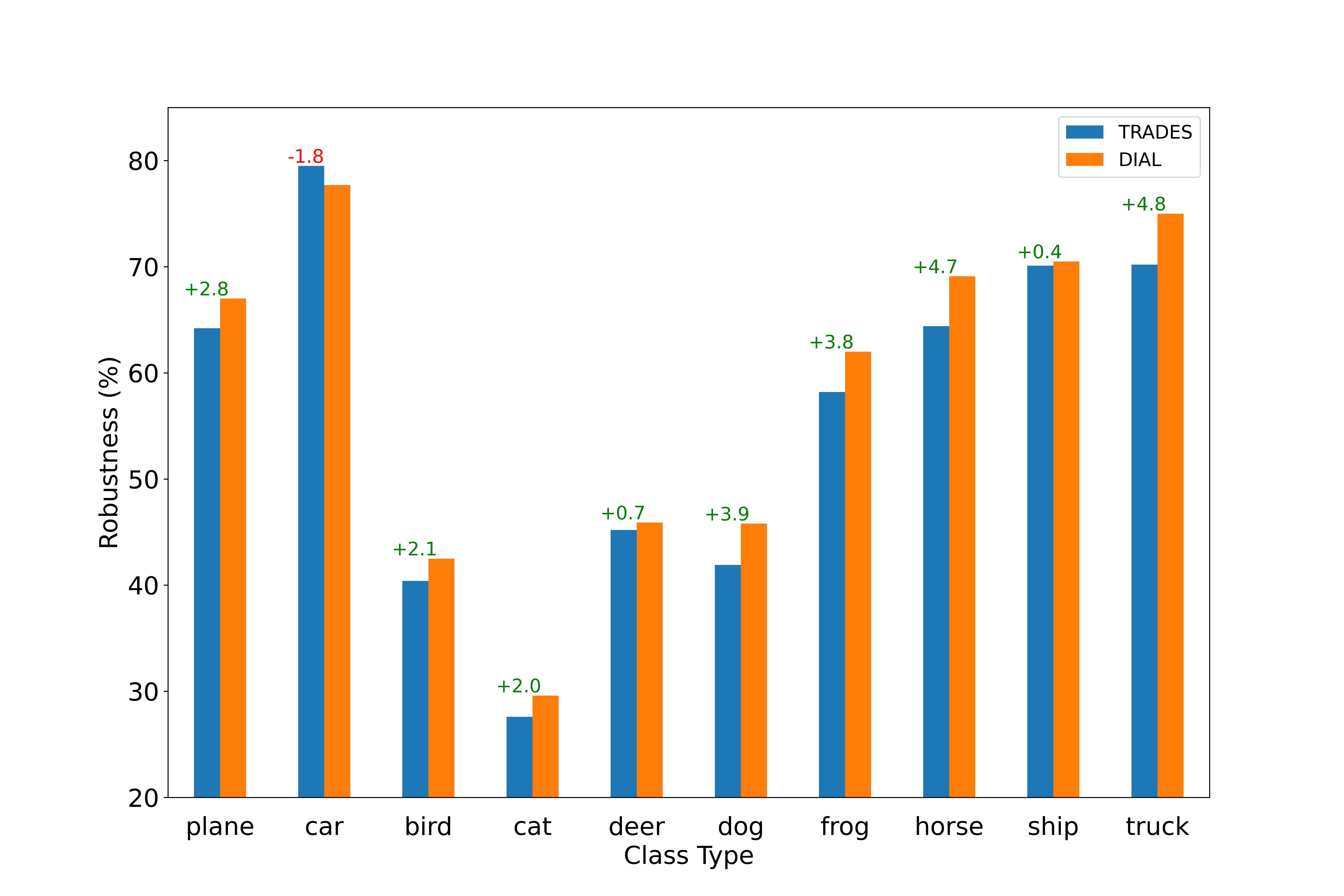}
  \caption{$\DIAL_{\kl}$ and TRADES Robustness (\%) for each class on CIFAR-10. Adversarial examples are generated using PGD-20. Our method manages to improve robustness over TRADES on 9 out of 10 classes. Green annotation presented the difference percentage improvement.}
  \label{trades-dial-adv}
\end{figure}

\begin{figure}[!ht]
  \centering
  \includegraphics[width=0.8\textwidth]{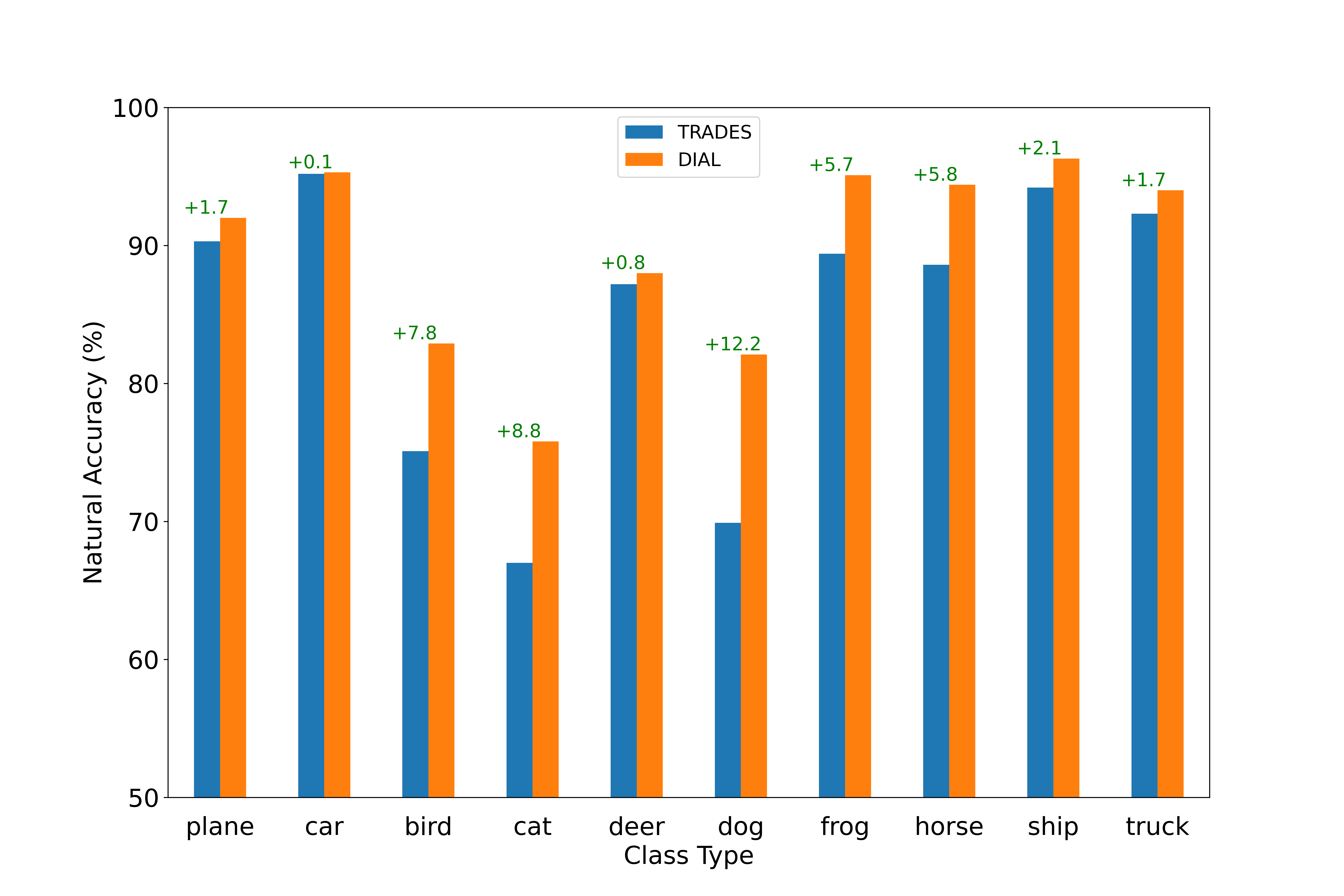}
  \caption{$\DIAL_{\ce}$ and TRADES natural accuracy (\%) for each class on CIFAR-10. Our method manages to improve natural accuracy on all 10 classes. Green annotation presented the difference percentage improvement.}
  \label{trades-dial-clean}
\end{figure}

Additionally, we present two additional results and visualizations that can show our performance difference is solid.
First, in Figure \ref{tsne-feat} we present a 2d T-SNE plot on the \textbf{features} layer (unlike the logits layer , which is more related to the quantitative results), to further demonstrate that our method indeed learns a domain invariant feature representation better than the other methods, and compared it to TRADES.
Second, we wish to demonstrate that our performance gain is not a due to specific aspects of the domain (e.g., improvement only on specific classes). To do so, we visualize in figures \ref{trades-dial-adv}, \ref{trades-dial-clean} the classification improvement obtained by our method on each class in CIFAR-10. As can be seen, on the natural examples, our method improved the accuracy on all classes, and on the adversarial examples, our method improves robustness on 9 out of 10 classes. This further demonstrates the generalization of our approach.

\end{document}